\pdfoutput=1

\documentclass[11pt]{article}

\usepackage{ACL2023}

\usepackage[switch]{lineno} 

\usepackage{flushend}

\usepackage{times}
\usepackage{graphicx}
\usepackage{latexsym}
\usepackage{multirow, makecell}
\usepackage{todonotes}
\usepackage{amsmath}
\usepackage{cleveref}
\usepackage{pifont}
\usepackage{pifont}
\usepackage{caption}

\usepackage{microtype}
\usepackage{subfigure}
\usepackage{booktabs} 
\usepackage{xcolor}
\usepackage{setspace}
\usepackage{multicol}
\usepackage{threeparttable}
\usepackage{tabulary}
\usepackage{colortbl}
\usepackage{ragged2e}
\usepackage{amsmath,bm}
\usepackage{adjustbox}

\usepackage{array}
\newcolumntype{x}[1]{>{\centering\arraybackslash\hspace{0pt}}p{#1}}
\newcolumntype{M}[1]{>{\centering\arraybackslash}m{#1}}

\usepackage{enumitem}
\usepackage{soul}

\newcommand{\hlc}[2][yellow]{{%
    \colorlet{foo}{#1}%
    \sethlcolor{foo}\hl{#2}}%
}

\usepackage{color}

\usepackage[T1]{fontenc}
\usepackage[utf8]{inputenc}

\usepackage{microtype}
\usepackage{colortbl}

\title{MatSci-NLP: Evaluating Scientific Language Models on Materials Science Language Tasks Using Text-to-Schema Modeling}

\definecolor{x3}{RGB}{255,230,204}
\definecolor{x4}{RGB}{255, 255, 255}

\definecolor{x0}{RGB}{226,109,16}
\definecolor{x1}{RGB}{255,181,122}

\definecolor{r1}{HTML}{f7a889}
\definecolor{r2}{HTML}{f6bfa6}
\definecolor{r3}{HTML}{edd1c2}
\definecolor{r4}{HTML}{dddcdc}
\definecolor{r5}{HTML}{c9d7f0}
\definecolor{r6}{HTML}{b2ccfb}
\definecolor{r7}{HTML}{9abbff}

\newcommand{\schema}{Task-Schema }
\newcommand{\benchmark}{MatSci-NLP }

\author{
Yu Song\textsuperscript{\rm 1,}\begin{NoHyper}\thanks{\;\; Equal contribution.}\end{NoHyper}\quad 
Santiago Miret\textsuperscript{\rm 2,}\footnotemark[1]\, \quad 
Bang Liu\textsuperscript{\rm 1,}\begin{NoHyper}\thanks{\;\; Corresponding author. Canada CIFAR AI Chair.}\end{NoHyper} \vspace{0.4em} \\
$^1$University of Montreal / Mila - Quebec AI,\, $^2$Intel Labs\\
\texttt{\{yu.song, bang.liu\}@umontreal.ca}\\
\texttt{\{santiago.miret\}@intel.com}\\
}

\begin{document}

\maketitle

\begin{abstract}
We present MatSci-NLP, a natural language benchmark for evaluating the performance of natural language processing (NLP) models on materials science text. We construct the benchmark from publicly available materials science text data to encompass seven different NLP tasks, including conventional NLP tasks like named entity recognition and relation classification, as well as NLP tasks specific to materials science, such as synthesis action retrieval which relates to creating synthesis procedures for materials. 
We study various BERT-based models pretrained on different scientific text corpora on MatSci-NLP to understand the impact of pretraining strategies on understanding materials science text. 
Given the scarcity of high-quality annotated data in the materials science domain, we perform our fine-tuning experiments with limited training data to encourage the generalize across MatSci-NLP tasks.
Our experiments in this low-resource training setting show that language models pretrained on scientific text outperform BERT trained on general text. 
MatBERT, a model pretrained specifically on materials science journals, generally performs best for most tasks. 
Moreover, we propose a unified text-to-schema for multitask learning on \benchmark and compare its performance with traditional fine-tuning methods. In our analysis of different training methods, we find that our proposed text-to-schema methods inspired by question-answering consistently outperform single and multitask NLP fine-tuning methods. The code and datasets are publicly available\footnote{\url{https://github.com/BangLab-UdeM-Mila/NLP4MatSci-ACL23}}.
\end{abstract}

\section{Introduction} \label{sec:intro}
Materials science comprises an interdisciplinary scientific field that studies the behavior, properties and applications of matter that make up materials systems. As such, materials science often requires deep understanding of a diverse set of scientific disciplines to meaningfully further the state of the art. This interdisciplinary nature, along with the great technological impact of materials advances and growing research work at the intersection of machine learning and materials science \citep{ai4mat, pilania2021machine, choudhary2022recent}, makes the challenge of developing and evaluating natural language processing (NLP) models on materials science text both interesting and exacting.

The vast amount of materials science knowledge stored in textual format, such as journal articles, patents and technical reports, creates a tremendous opportunity to develop and build NLP tools to create and understand advanced materials. These tools could in turn enable faster discovery, synthesis and deployment of new materials into a wide variety of application, including clean energy, sustainable manufacturing and devices.

Understanding, processing, and training language models for scientific text presents distinctive challenges that have given rise to the creation of specialized models and techniques that we review in \Cref{sec:background}. Additionally, evaluating models on scientific language understanding tasks, especially in materials science, often remains a laborious task given the shortness of high-quality annotated data and the lack of broad model benchmarks. As such, NLP research applied to materials science remains in the early stages with a plethora of ongoing research efforts focused on dataset creation, model training and domain specific applications.

The broader goal of this work is to enable the development of pertinent language models that can be applied to further the discovery of new material systems, and thereby get a better sense of how well language models understand the properties and behavior of existing and new materials.
As such, we propose MatSci-NLP, a benchmark of various NLP tasks spanning many applications in the materials science domain described in \Cref{sec:meta-dataset}. We utilize this benchmark to analyze the performance of various BERT-based models for \benchmark tasks under distinct textual input schemas described in \Cref{sec:schema}. 
Concretely, through this work we make the following research contributions: 
\begin{itemize}
    \item \textbf{MatSci-NLP Benchmark:} We construct the first broad benchmark for NLP in the materials science domain, spanning several different NLP tasks and materials applications. The benchmark contents are described in \Cref{sec:meta-dataset} with a general summary and data sources provided in \Cref{table:tasks}. The processed datasets and code will be released after acceptance of the paper for reproducibility.
    \item \textbf{Text-to-Schema Multitasking:} We develop a set of textual input schemas inspired by question-answering settings for fine-tuning language models. We analyze the models' performance on \benchmark across those settings and conventional single and multitask fine-tuning methods. 
    In conjunction with this analysis, we propose a new \schema input format for joint multitask training that increases task performance for all fine-tuned language models.
    \item \textbf{MatSci-NLP Analysis:} We analyze the performance of various BERT-based models pretrained on different scientific and non-scientific text corpora on the MatSci-NLP benchmark. This analysis help us better understand how different pretraining strategies affect downstream tasks and find that MatBERT \citep{walker2021impact}, a BERT model trained on materials science journals, generally performs best reinforcing the importance of curating high-quality pretraining corpora.
\end{itemize}

We centered our MatSci-MLP analysis on exploring the following questions:
    \begin{enumerate}
    [label=Q\arabic*]
        \item\label{q1} \emph{How does in-domain pretraining of language models affect the downstream performance on MatSci-NLP tasks?} We investigate the performance of various models pretrained on different kinds of domain-specific text including materials science, general science and general language (BERT \citep{devlin2018bert}). We find that MatBERT generally performs best and that language models pretrained on diverse scientific texts outperform a general language BERT. Interestingly, SciBERT \citep{beltagy2019scibert} often outperforms materials science language models, such as MatSciBERT \citep{gupta2022matscibert} and BatteryBERT \citep{huang2022batterybert}.
        \item\label{q2} \emph{How do in-context data schema and multitasking affect the learning efficiency in low-resource training settings?} We investigate how several input schemas shown in \Cref{fig:matsci-nlp-example} that contain different kinds of information affect various domain-specific language models and propose a new \textit{Task-Schema} method. Our experiments show that our proposed Task-Schema method mostly performs best across all models and that question-answering inspired schema outperform single task and multitask fine-tuning settings. 
    \end{enumerate}

\section{Background} \label{sec:background}

The advent of powerful NLP models has enabled the analysis and generation of text-based data across a variety of domains. BERT \citep{devlin2018bert} was one of the first large-scale transformer-based models to substantially advance the state-of-the-art by training on large amounts of unlabeled text data in a self-supervised way. The pretraining procedure was followed by task-specific fine-tuning, leading to impressive results on a variety of NLP task, such as named entity recognition (NER), question and answering (QA), and relation classification \citep{hakala2019biomedical, qu2019bert, wu2019enriching}. 
A significant collection of large language models spanning millions to billions of parameters followed the success of BERT adopting a similar approach of pretraining on vast corpora of text with task-specific fine-tuning to push the state-of-the-art for in natural language processing and understanding \citep{raffel2020exploring,brown2020language, scao2022bloom}. 

\subsection{Scientific Language Models}
The success of large language models on general text motivated the development of domain-specific language models pretrained on custom text data, including text in the scientific domain: SciBERT \citep{beltagy2019scibert}, ScholarBERT \citep{hong2022scholarbert} and Galactica \citep{taylor2022galactica} are pretrained on general corpus of scientific articles; BioBERT \citep{lee2020biobert}, PubMedBERT \citep{gu2021domain}, BioMegatron \citep{shin2020biomegatron} and Sci-Five \citep{phan2021scifive} are pretrained on various kinds of biomedical corpora; MatBERT \citep{walker2021impact}, MatSciBERT \citep{gupta2022matscibert} are pretrained on materials science specific corpora; and BatteryBERT \citep{huang2022batterybert} is pretrained on a corpus focused on batteries. 
Concurrently, several domain-specific NLP benchmarks were established to assess language model performance on domain-specific tasks, such as QASPER \citep{dasigi2021dataset} and BLURB \citep{gu2021domain} in the scientific domain, as well as PubMedQA \citep{jin2019pubmedqa}, BioASQ \citep{balikas2015bioasq}, and Biomedical Language Understanding Evaluation (BLUE) \citep{peng2019transfer} in the biomedical domain.

\subsection{NLP in Materials Science}

The availability of openly accessible, high-quality corpora of materials science text data remains highly restricted in large part because data from peer-reviewed journals and scientific documents is usually subject to copyright restrictions, while open-domain data is often only available in difficult-to-process PDF formats \citep{olivetti2020data, kononova2021opportunities}. Moreover, specialized scientific text, such as materials synthesis procedures containing chemical formulas and reaction notation, require advanced data mining techniques for effective processing \citep{kuniyoshi2020annotating, wang2022dataset}. 
Given the specificity, complexity, and diversity of specialized language in scientific text, effective extraction and processing remain an active area of research with the goal of building relevant and sizeable text corpora for pretraining scientific language models \citep{kononova2021opportunities}.

Nonetheless, materials science-specific language models, including MatBERT \citep{walker2021impact},  MatSciBERT \citep{gupta2022matscibert}, and BatteryBERT \citep{huang2022batterybert}, have been trained on custom-built pretraining dataset curated by different academic research groups. The pretrained models and some of the associated fine-tuning data have been released to the public and have enabled further research, including this work. 

The nature of NLP research in materials science to date has also been highly fragmented with many research works focusing on distinct tasks motivated by a given application or methodology. Common ideas among many works include the prediction and construction of synthesis routes for a variety of materials \citep{mahbub2020text, karpovich2021inorganic, kim2020inorganic}, as well as the creation of novel materials for a given application \citep{huang2022batterybert, georgescu2021database, jensen2021discovering}, both of which relate broader challenges in the field of materials science. 

\begin{figure*}[h]
\centering
    \includegraphics[width=1\linewidth]{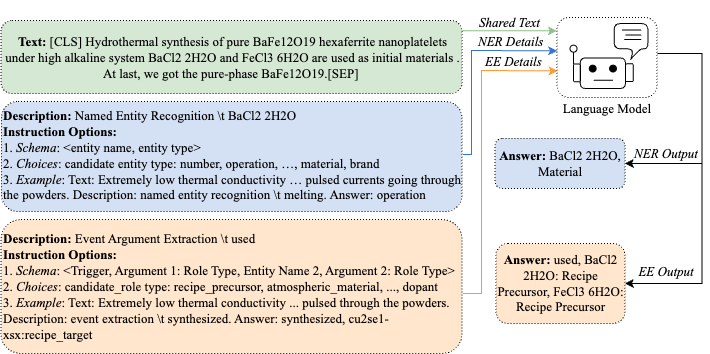}
    \caption{Example of different question-answering inspired textual input schemas (\schema, Potential Choices, Example) applied on MatSci-NLP. The input of the language model includes the shared text (green) along with relevant task details (blue for NER and orange for event extraction). The shared text can contain relevant information for multiple tasks and be part of the language model input multiple times.}
    \label{fig:matsci-nlp-example}
\end{figure*}


\section{MatSci-NLP Benchmark} \label{sec:meta-dataset}
Through the creation of MatSci-NLP, we aim to bring together some of the fragmented data across multiple research works for a wide-ranging materials science NLP benchmark. As described in \Cref{sec:background}, the availability of sizeable, high-quality and diverse datasets remain a major obstacle in applying modern NLP to advance materials science in meaningful ways. This is primarily driven by a high cost of data labeling and the heterogeneous nature of materials science. Given those challenges, we created \benchmark by unifying various publicly available, high-quality, smaller-scale datasets to form a benchmark for fine-tuning and evaluating modern NLP models for materials science applications.  \benchmark consists of seven NLP tasks shown in \Cref{table:tasks}, spanning a wide range of materials categories including fuel cells \citep{friedrich2020sofc}, glasses \citep{venugopal2021looking}, inorganic materials \citep{weston2019named, matscire2022github}, superconductors \citep{yamaguchi2020sc}, and synthesis procedures pertaining to various kinds of materials \citep{mysore2019materials, wang2022ulsa}. Some tasks in \benchmark had multiple source components, meaning that the data was curated from multiple datasets (e.g. NER), while many were obtained from a single source dataset.


\begin{table}[h!]
\begin{spacing}{1.07}
\centering

\vspace{-2.5mm}
\begin{adjustbox}{max width=1\linewidth}
\small
    \begin{tabular}{l|c|c}
        \toprule
        \multicolumn{1}{c}{\bf{Task}} & \multicolumn{1}{c}{\makecell{\bf{Size} \\ (\# Samples)}} & \makecell{\bf{Meta-Dataset} \\ \bf{Components}}  \\
        \midrule
        \multicolumn{1}{c}{\makecell{Named Entity \\ Recognition}} & \multicolumn{1}{c}{\makecell{112,191}} & \multicolumn{1}{c}{4} \\ 
        \midrule
        \multicolumn{1}{c}{\makecell{Relation \\ Classification}} & \multicolumn{1}{c}{\makecell{25,674}} & \multicolumn{1}{c}{3} \\
        \midrule
        \multicolumn{1}{c}{\makecell{Event Argument \\ Extraction}} & \multicolumn{1}{c}{\makecell{6,566}} & \multicolumn{1}{c}{2} \\
        \midrule
        \multicolumn{1}{c}{\makecell{Paragraph \\ Classification}} & \multicolumn{1}{c}{\makecell{1,500}} & \multicolumn{1}{c}{1} \\
        \midrule
        \multicolumn{1}{c}{\makecell{Synthesis \\ Action Retrieval}} & \multicolumn{1}{c}{\makecell{5,547}} & \multicolumn{1}{c}{1} \\
        \midrule
        \multicolumn{1}{c}{\makecell{Sentence \\ Classification}} & \multicolumn{1}{c}{\makecell{9,466}} & \multicolumn{1}{c}{1} \\
        \midrule
        \multicolumn{1}{c}{\makecell{Slot Filling}} & \multicolumn{1}{c}{\makecell{8,253}} & \multicolumn{1}{c}{1} \\
               
        \bottomrule
    \end{tabular}
\end{adjustbox}
\end{spacing}
\vspace{-2mm}
\caption{Collection of NLP tasks in the meta-dataset of the MatSci-NLP Benchmark drawn from \citet{weston2019named,friedrich2020sofc,mysore2019materials,yamaguchi2020sc,venugopal2021looking,wang2022ulsa,matscire2022github}.}
\label{table:tasks}
\end{table}


The data in \benchmark adheres to a standard JSON-based data format with each of the samples containing relevant text, task definitions, and annotations. 
These can in turn be refactored into different input schemas, such as the ones shown in \Cref{fig:matsci-nlp-example} consisting of 1) \emph{Input}: primary text jointly with task descriptions and instructions, and 2) \emph{Output}: query and label, which we perform in our text-to-schema modeling described in \Cref{sec:schema}. 
Next, we describe the tasks in MatSci-NLP in greater detail:

\begin{itemize}
    \item \textbf{Named Entity Recognition~(NER):} The NER task requires models to extract summary-level information from materials science text and recognize entities including materials, descriptors, material properties, and applications amongst others. The NER task predicts the best entity type label for a given text span $s_i$ with a non-entity span containing a “null” label. MatSci-NLP contains NER task data adapted from \citet{weston2019named, friedrich2020sofc, mysore2019materials, yamaguchi2020sc}.
    \item \textbf{Relation Classification: } In the relation classification task, the model predicts the most relevant relation type for a given span pair $(s_i, s_j)$. MatSci-NLP contains relation classification task data adapted from \citet{mysore2019materials, yamaguchi2020sc, matscire2022github}. 
    \item \textbf{Event Argument Extraction: } The event argument extraction task involves extracting event arguments and relevant argument roles. As there may be more than a single event for a given text, we specify event triggers and require the language model to extract corresponding arguments and their roles. MatSci-NLP contains event argument extraction task data adapted from \citet{mysore2019materials, yamaguchi2020sc}.
    \item \textbf{Paragraph Classification: } In the paragraph classification task adapted from \citet{venugopal2021looking}, the model determines whether a given paragraph pertains to glass science.
    \item \textbf{Synthesis Action Retrieval~(SAR): } SAR is a materials science domain-specific task that defines eight action terms that unambiguously identify a type of synthesis action to describe a synthesis procedure. MatSci-NLP adapts SAR data from \citet{wang2022ulsa} to ask language models to classify word tokens into pre-defined action categories.
    \item \textbf{Sentence Classification: } In the sentence classification task, models identify sentences that describe relevant experimental facts based on data adapted from \citet{friedrich2020sofc}.
    \item \textbf{Slot Filling: } In the slot-filling task, models extract slot fillers from particular sentences based on a predefined set of semantically meaningful entities. In the task data adapted from \citet{friedrich2020sofc}, each sentence describes a single experiment frame for which the model predicts the slots in that frame.
\end{itemize}

The tasks contained in \benchmark were selected based on publicly available, high-quality annotated materials science textual data, as well as their relevance to applying NLP tools to materials science. Conventional NLP tasks (NER, Relation Classification, Event Argument Extraction, Paragraph Classification, Sentence Classification) enable materials science researchers to better process and understand relevant textual data. Domain specific tasks (SAR, Slot Filling) enable materials science research to solve concrete challenges, such as finding materials synthesis procedures and real-world experimental planning. In the future, we aim to augment to current set of tasks with additional data and introduce novel tasks that address materials science specific challenges with NLP tools. 


\section{Unified Text-to-Schema Language Modeling} \label{sec:schema}

\begin{figure}[h]
\centering
    \includegraphics[width=0.9\linewidth]{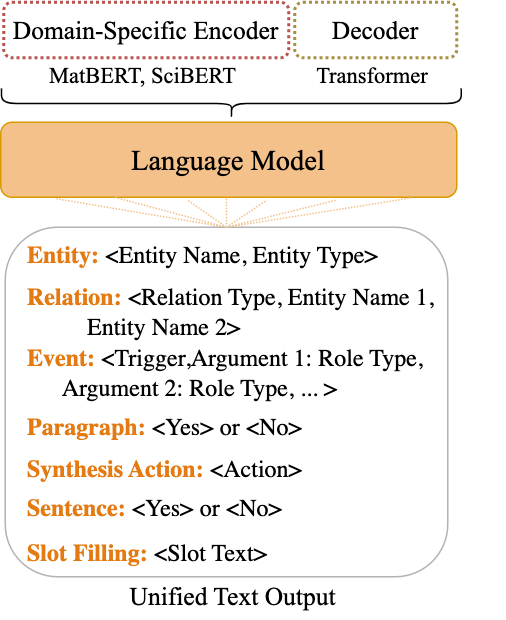}
    \caption{Unified text-to-schema method for MatSci-NLP text understanding applied across the seven tasks. The language model includes a domain specific encoder, which can be exchanged in a modular manner, as well as a general language pretrained transformer decoder.}
    \label{fig:matsci-nlp-method}
\end{figure}

As shown in \Cref{fig:matsci-nlp-example}, a given piece of text can include multiple labels across different tasks. 
Given this multitask nature of the MatSci-NLP benchmark, we propose a new and unified \textit{\schema} multitask modeling method illustrated in \Cref{fig:matsci-nlp-method} that covers all the tasks in the MatSci-NLP dataset. 
Our approach centers on a unified text-to-schema modeling approach that can predict multiple tasks simultaneously through a unified format. The underlying language model architecture is made up of modular components, including a domain-specific encoder model (e.g. MatBERT, MatSciBERT, SciBERT), and a generic transformer-based decoder, each of which can be easily exchanged with different pretrained domain-specific NLP models. We fine-tune these pretrained language models and the decoder with collected tasks in \benchmark using the procedure described in \Cref{sec:decoding}.

The unified text-to-schema provides a more structured format to training and evaluating language model outputs compared to seq2seq and text-to-text approaches \citep{raffel2020exploring, luong2015multi}. This is particularly helpful for the tasks in \benchmark given that many tasks can be reformulated as classification problems. NER and Slot Filling, for example, are classifications at the token-level, while event arguments extraction entails the classification of roles of certain arguments. Without a predefined schema, the model relies entirely on unstructured natural language to provide the answer in a seq2seq manner, which significantly increases the complexity of the task and also makes it harder to evaluate performance. 
The structure imposed by text-to-schema method also simplifies complex tasks, such as event extraction, by enabling the language model to leverage the structure of the schema to predict the correct answer. We utilize the structure of the schema in decoding and evaluating the output of the language models, as described further in \Cref{sec:decoding} in greater detail. 

Moreover, our unified text-to-schema approach alleviates error propagation commonly found in multitask scenarios \citep{van2022joint, lu2021text2event}, enables knowledge sharing across multiple tasks and encourages the fine-tuned language model to generalize across a broader set of text-based instruction scenarios. This is supported by our results shown in \Cref{sec:results-q2} showing text-to-schema outperforming conventional methods.

\subsection{Language Model Formulation} \label{sec:formulation}
The general purpose of our model is to achieve multitask learning by a mapping function ($f$) between input ($x$), output ($y$), and schema ($s$), i.e., $f(x,s)= y$. Due to the multitasking nature of our setting, both inputs and outputs can originate from different tasks n, i.e. $x = [x_t1, x_t2, ... x_tn]$ and $y = [y_t1, y_t2, ... y_tn]$, all of which fit under a common schema ($s$).
Given the presence of domain-specific materials science language, our model architecture includes a domain-specific BERT encoder and a transformer decoder. All BERT encoders and transformer decoders share the same general architecture, which relies on a self-attention mechanism: Given an input sequence of length $N$, we compute a set of attention scores, $A = \rm{softmax}(QT^K/(\sqrt{d_k}))$. Next, we compute the weighted sum of the value vectors, $O = AV$, where $Q$, $K$, and $V$ are the query, key, and value matrices, and $d_k$ is the dimensionality of the key vectors.

Additionally, the transformer based decoder differ from the domain specific encoder by:
1) Applying masking based on the schema applied to ensure that it does not attend to future positions in the output sequence. 
2) Applying both self-attention and encoder-decoder attention to compute attention scores that weigh the importance of different parts of the output sequence and input sequence. The output of the self-attention mechanism ($O_1$) and the output of the encoder-decoder attention mechanism ($O_2$) are concatenated and linearly transformed to obtain a new hidden state, $H = \rm{tanh}(W_o[O_1;O_2] + b_o)$ with $W_o$ and $b_o$ being the weight and biases respectively. The model then applies a \emph{softmax} to $H$ to generate the next element in the output sequence $P = \rm{softmax}(W_pH + b_p)$ , where $P$ is a probability distribution over the output vocabulary.

\subsection{Text-To-Schema Modeling}

As shown in \Cref{fig:matsci-nlp-example}, our schema structures the text data based on four general components: text, description, instruction options, and the predefined answer schema. 
\begin{itemize}
    \item \textbf{Text} specifies raw text from the literature that is given as input to the language model.
    \item \textbf{Description} describes the task for a given text according to a predefined schema containing the task name and the task arguments.
    \item \textbf{Instruction Options} contains the core explanation related to the task with emphasis on three different types: 1) Potential choices of answers; 2) Example of an input/output pair corresponding to the task; 3) \schema: our predefined answer schema illustrated in ~\Cref{fig:matsci-nlp-method}.
    \item \textbf{Answer} describes the correct label of each task formatted as a predefined answer schema that can be automatically generated based on the data structure of the task.
\end{itemize}

\subsection{Language Decoding \& Evaluation} \label{sec:decoding}

Evaluating the performance of the language model on MatSci-NLP requires determining if the text generated by the decoder is valid and meaningful in the context of a given task. 
To ensure consistency in evaluation, we apply a constrained decoding procedure consisting of two steps: 1) Filtering out invalid answers through the predefined answer schema shown in \Cref{fig:matsci-nlp-method} based on the structure of the model's output; 2) Match the model's prediction with the most similar valid class given by the annotation for the particular task. For example, if for the NER task shown in \Cref{fig:matsci-nlp-example} the model's predicted token is ``BaCl2 2H2O materials'', it will be matched with the NER label of ``material'', which is then used as the final prediction for computing losses and evaluating performance.
This approach essentially reformulates each task as a classification problem where the classes are provided based on the labels from the tasks in MatSci-NLP. We then apply a cross-entropy loss for model fine-tuning based on the matched label from the model output. The matching procedure simplifies the language modeling challenge by not requiring an exact match of the predicted tokens with the task labels. This in turns leads to a more comprehensible signal in the fine-tuning loss function.


\begin{table*}[h]
\begin{spacing}{1.07}
\centering

\vspace{-2.5mm}
\begin{adjustbox}{max width=1\linewidth}
    \begin{tabular}{c|c|c|c|c|c|c|c|c}
        \toprule
        \bf{NLP Model} & \multicolumn{1}{c}{\makecell{\bf{Named Entity} \\ \bf{Recognition}}} & \multicolumn{1}{c}{\makecell{\bf{Relation} \\ \bf{Classification}}} 
         & \multicolumn{1}{c}{\makecell{\bf{Event Argument} \\ \bf{Extraction}}}
         & \multicolumn{1}{c}{\makecell{\bf{Paragraph} \\ \bf{Classification}}}
         & \multicolumn{1}{c}{\makecell{\bf{Synthesis} \\ \bf{Action Retrieval}}}
         & \multicolumn{1}{c}{\makecell{\bf{Sentence} \\ \bf{Classification}}}
         & \multicolumn{1}{c}{\makecell{\bf{Slot} \\ \bf{Filling}}}
         & \multicolumn{1}{c}{\makecell{\bf{Overall} \\ \bf{(All Tasks)} }}\\
        \midrule
        \makecell{MatSciBERT \\ \citep{gupta2022matscibert}} & 
        \multicolumn{1}{c}{\cellcolor{x3} \makecell{0.707$_{\pm 0.076}$ \\ 0.470$_{\pm 0.092}$}} & 
        \multicolumn{1}{c}{\cellcolor{x3} \makecell{0.791$_{\pm 0.046}$ \\ 0.507$_{\pm 0.073}$}} & 
        \multicolumn{1}{c}{\cellcolor{x3} \makecell{0.436$_{\pm 0.066}$ \\ 0.251$_{\pm 0.075}$}} & 
        \multicolumn{1}{c}{\cellcolor{x3}  \makecell{0.719$_{\pm 0.116}$ \\ 0.623$_{\pm 0.183}$}} & 
        \multicolumn{1}{c}{\cellcolor{x3} \makecell{0.692$_{\pm 0.179}$ \\ 0.484$_{\pm 0.254}$}} & 
        \multicolumn{1}{c}{\cellcolor{x3} \makecell{0.914$_{\pm 0.008}$ \\ 0.660$_{\pm 0.079}$}} & 
        \multicolumn{1}{c}{\cellcolor{x4} \makecell{0.436$_{\pm 0.142}$ \\ 0.194$_{\pm 0.062}$}} & 
        \multicolumn{1}{c}{\cellcolor{x3} \makecell{0.671$_{\pm 0.060}$ \\ 0.456$_{\pm 0.042}$}}\\ 
        \midrule
        \makecell{MatBERT \\ \citep{walker2021impact}} &  
        \multicolumn{1}{c}{\cellcolor{x1} \makecell{0.875$_{\pm 0.015}$ \\ 0.630$_{\pm 0.047}$}} & 
        \multicolumn{1}{c}{\cellcolor{x3} \makecell{0.804$_{\pm 0.071}$ \\ 0.513$_{\pm 0.138}$}} & 
        \multicolumn{1}{c}{\cellcolor{x3} \makecell{0.451$_{\pm 0.091}$ \\ 0.288$_{\pm 0.066}$}} & 
        \multicolumn{1}{c}{\cellcolor{x1} \makecell{0.756$_{\pm 0.073}$ \\ 0.691$_{\pm 0.188}$}} & 
        \multicolumn{1}{c}{\cellcolor{x1} \makecell{0.717$_{\pm 0.040}$ \\ 0.549$_{\pm 0.091}$}} & 
        \multicolumn{1}{c}{\cellcolor{x4} \makecell{0.909$_{\pm 0.009}$ \\ 0.614$_{\pm 0.134}$}} & 
        \multicolumn{1}{c}{\cellcolor{x1} \makecell{0.548$_{\pm 0.058}$ \\ 0.273$_{\pm 0.051}$}} &
        \multicolumn{1}{c}{\cellcolor{x1} \makecell{0.722$_{\pm 0.023}$ \\ 0.517$_{\pm 0.041}$}} \\ 
        \midrule
        \makecell{BatteryBERT \\ \citep{huang2022batterybert}} &  
        \multicolumn{1}{c}{\cellcolor{x3} \makecell{0.786$_{\pm 0.113}$ \\ 0.472$_{\pm 0.150}$}} & 
        \multicolumn{1}{c}{\cellcolor{x3} \makecell{0.801$_{\pm 0.081}$ \\ 0.466$_{\pm 0.111}$}} & 
        \multicolumn{1}{c}{\cellcolor{x3} \makecell{0.457$_{\pm 0.024}$ \\ 0.277$_{\pm 0.034}$}} & 
        \multicolumn{1}{c}{\cellcolor{x4} \makecell{0.633$_{\pm 0.075}$ \\ 0.610$_{\pm 0.046}$}} & 
        \multicolumn{1}{c}{\cellcolor{x4} \makecell{0.614$_{\pm 0.128}$ \\ 0.419$_{\pm 0.149}$}} & 
        \multicolumn{1}{c}{\cellcolor{x3} \makecell{0.912$_{\pm 0.015}$ \\ 0.684$_{\pm 0.095}$}} & 
        \multicolumn{1}{c}{\cellcolor{x3} \makecell{0.520$_{\pm 0.057}$ \\ 0.224$_{\pm 0.073}$}} &
        \multicolumn{1}{c}{\cellcolor{x3} \makecell{0.663$_{\pm 0.038}$ \\ 0.456$_{\pm 0.048}$}} \\ 
        \midrule
        \makecell{SciBERT \\ \citep{beltagy2019scibert}} & 
        \multicolumn{1}{c}{\cellcolor{x3} \makecell{0.734$_{\pm 0.079}$ \\ 0.497$_{\pm 0.091}$}} & 
        \multicolumn{1}{c}{\cellcolor{x1} \makecell{0.819$_{\pm 0.067}$ \\ 0.545$_{\pm 0.119}$}} & 
        \multicolumn{1}{c}{\cellcolor{x3} \makecell{0.451$_{\pm 0.077}$ \\ 0.276$_{\pm 0.080}$}} & 
        \multicolumn{1}{c}{\cellcolor{x3} \makecell{0.696$_{\pm 0.094}$ \\ 0.546$_{\pm 0.243}$}} & 
        \multicolumn{1}{c}{\cellcolor{x3} \makecell{0.701$_{\pm 0.138}$ \\ 0.516$_{\pm 0.217}$}} & 
        \multicolumn{1}{c}{\cellcolor{x3} \makecell{0.911$_{\pm 0.017}$ \\ 0.617$_{\pm 0.143}$}}& 
        \multicolumn{1}{c}{\cellcolor{x4} \makecell{0.481$_{\pm 0.144}$ \\ 0.224$_{\pm 0.010}$}}& 
        \multicolumn{1}{c}{\cellcolor{x3} \makecell{0.685$_{\pm 0.056}$ \\ 0.460$_{\pm 0.044}$}}\\ 
        \midrule
        \makecell{ScholarBERT \\ \citep{hong2022scholarbert}} & 
        \multicolumn{1}{c}{\cellcolor{x4} \makecell{0.168$_{\pm 0.067}$ \\ 0.101$_{\pm 0.034}$}} & 
        \multicolumn{1}{c}{\cellcolor{x4} \makecell{0.428$_{\pm 0.148}$ \\ 0.274$_{\pm 0.110}$}} & 
        \multicolumn{1}{c}{\cellcolor{x1} \makecell{0.489$_{\pm 0.083}$ \\ 0.356$_{\pm 0.109}$}} & 
        \multicolumn{1}{c}{\cellcolor{x4} \makecell{0.663$_{\pm 0.032}$ \\ 0.433$_{\pm 0.122}$}} & 
        \multicolumn{1}{c}{\cellcolor{x4} \makecell{0.322$_{\pm 0.260}$ \\ 0.178$_{\pm 0.051}$}} & 
        \multicolumn{1}{c}{\cellcolor{x4} \makecell{0.906$_{\pm 0.007}$ \\ 0.478$_{\pm 0.008}$}}& 
        \multicolumn{1}{c}{\cellcolor{x4} \makecell{0.296$_{\pm 0.085}$ \\ 0.109$_{\pm 0.044}$}}& 
        \multicolumn{1}{c}{\cellcolor{x4} \makecell{0.468$_{\pm 0.028}$ \\ 0.276$_{\pm 0.024}$}}\\ 
        \midrule
        \makecell{BioBERT \\ \citep{shoyabiobert}} & 
        \multicolumn{1}{c}{\cellcolor{x3} \makecell{0.715$_{\pm 0.031}$ \\ 0.459$_{\pm 0.055}$}} & 
        \multicolumn{1}{c}{\cellcolor{x3} \makecell{0.797$_{\pm 0.092}$ \\ 0.465$_{\pm 0.134}$}} & 
        \multicolumn{1}{c}{\cellcolor{x3} \makecell{0.488$_{\pm 0.036}$ \\ 0.274$_{\pm 0.049}$}} & 
        \multicolumn{1}{c}{\cellcolor{x3} \makecell{0.675$_{\pm 0.144}$ \\ 0.578$_{\pm 0.102}$}} & 
        \multicolumn{1}{c}{\cellcolor{x4} \makecell{0.647$_{\pm 0.140}$ \\ 0.446$_{\pm 0.231}$}} & 
        \multicolumn{1}{c}{\cellcolor{x1} \makecell{0.915$_{\pm 0.021}$ \\ 0.686$_{\pm 0.098}$}}& 
        \multicolumn{1}{c}{\cellcolor{x4} \makecell{0.452$_{\pm 0.114}$ \\ 0.191$_{\pm 0.045}$}}& 
        \multicolumn{1}{c}{\cellcolor{x3} \makecell{0.670$_{\pm 0.061}$ \\ 0.442$_{\pm 0.057}$}}\\ 
        \midrule
        \makecell{BERT \\ \citep{devlin2018bert}} & 
        \multicolumn{1}{c}{\makecell{0.657$_{\pm 0.077}$ \\ 0.461$_{\pm 0.058}$}} & 
        \multicolumn{1}{c}{\makecell{0.782$_{\pm 0.056}$ \\ 0.494$_{\pm 0.061}$}} & 
        \multicolumn{1}{c}{\makecell{0.418$_{\pm 0.053}$ \\ 0.225$_{\pm 0.091}$}} & 
        \multicolumn{1}{c}{\makecell{0.665$_{\pm 0.057}$ \\ 0.532$_{\pm 0.194}$}} & 
        \multicolumn{1}{c}{\makecell{0.656$_{\pm 0.099}$ \\ 0.515$_{\pm 0.067}$}} & 
        \multicolumn{1}{c}{\makecell{0.910$_{\pm 0.017}$ \\ 0.633$_{\pm 0.133}$}}& 
        \multicolumn{1}{c}{\makecell{0.520$_{\pm 0.019}$ \\ 0.257$_{\pm 0.022}$}} & 
        \multicolumn{1}{c}{\makecell{0.658$_{\pm 0.030}$ \\ 0.439$_{\pm 0.021}$}}\\ 
        \bottomrule
    \end{tabular}
\end{adjustbox}
\end{spacing}
\vspace{-2mm}
\caption{Low-resource fine-tuning results applying unified \schema setting for various BERT-based encoder models pretrained on different domain specific text data. For each model, the top line represents the micro-F1 score and the bottom line represents the macro-F1 score. We report the mean across 5 experiments with a confidence interval of two standard deviations. We denote the \hlc[x1]{best} performing encoder model and those that \hlc[x3]{outperform the general language BERT} according to the micro-f1 with orange shading with MatBERT and SciBERT performing best on most tasks and ScholarBERT and general language BERT generally performing worst.}
\label{table:low-resource-setting}
\end{table*}

\section{Evaluation and Results} \label{sec:results}

\begin{table*}[h]
\begin{spacing}{1.07}
\centering

\vspace{-2.5mm}
\begin{adjustbox}{max width=1\linewidth}
    \begin{tabular}{c|c|c|c|c|c|c|c}
        \toprule
        \bf{NLP Model} &
        \multicolumn{1}{c}{\makecell{\bf{Single Task}}} & 
        \multicolumn{1}{c}{\makecell{\bf{Single Task Prompt}}} & 
        \multicolumn{1}{c}{\makecell{\bf{MMOE}}} & 
        \multicolumn{1}{c}{\makecell{\bf{No Explanations}}} & 
        \multicolumn{1}{c}{\makecell{\bf{Potential} \bf{Choices}}} & 
        \multicolumn{1}{c}{\makecell{\bf{Examples}}} &
        \multicolumn{1}{c}{\makecell{\bf{\schema}}} \\
        \midrule
        \makecell{MatSciBERT \\ \citep{gupta2022matscibert}} & 
        \multicolumn{1}{c}{\cellcolor{r5} \makecell{0.501$_{\pm 0.057}$ \\ 0.320$_{\pm 0.078}$}} & 
        \multicolumn{1}{c}{\cellcolor{r6} \makecell{0.485$_{\pm 0.043}$ \\ 0.238$_{\pm 0.017}$}} & 
        \multicolumn{1}{c}{\cellcolor{r7} \makecell{0.457$_{\pm 0.021}$ \\ 0.228$_{\pm 0.038}$}} & 
        \multicolumn{1}{c}{\cellcolor{r4} \makecell{0.651$_{\pm 0.045}$ \\ 0.438$_{\pm 0.052}$}} & 
        \multicolumn{1}{c}{\cellcolor{r3} \makecell{0.670$_{\pm 0.036}$ \\ 0.435$_{\pm 0.061}$}} & 
        \multicolumn{1}{c}{\cellcolor{r1} \makecell{0.688$_{\pm 0.045}$ \\ 0.463$_{\pm 0.040}$}} & 
        \multicolumn{1}{c}{\cellcolor{r2} \makecell{0.671$_{\pm 0.060}$ \\ 0.456$_{\pm 0.042}$}} \\
        \midrule
        \makecell{MatBERT \\ \citep{walker2021impact}} &  
        \multicolumn{1}{c}{\cellcolor{r6} \makecell{0.537$_{\pm 0.036}$ \\ 0.330$_{\pm 0.063}$}} & 
        \multicolumn{1}{c}{\cellcolor{r7}\makecell{0.523$_{\pm 0.021}$ \\ 0.267$_{\pm 0.014}$}} & 
        \multicolumn{1}{c}{\cellcolor{r5} \makecell{0.557$_{\pm 0.010}$ \\ 0.301$_{\pm 0.006}$}} & 
        \multicolumn{1}{c}{\cellcolor{r2} \makecell{0.721$_{\pm 0.033}$ \\ 0.514$_{\pm 0.045}$}} & 
        \multicolumn{1}{c}{\cellcolor{r4} \makecell{0.699$_{\pm 0.020}$ \\ 0.478$_{\pm 0.032}$}} & 
        \multicolumn{1}{c}{\cellcolor{r3} \makecell{0.705$_{\pm 0.025}$ \\ 0.470$_{\pm 0.029}$}} & 
        \multicolumn{1}{c}{\cellcolor{r1} \makecell{0.722$_{\pm 0.023}$ \\ 0.517$_{\pm 0.041}$}} \\
        \midrule
        \makecell{BatteryBERT \\ \citep{huang2022batterybert}} &  
        \multicolumn{1}{c}{\cellcolor{r6} \makecell{0.469$_{\pm 0.050}$ \\ 0.288$_{\pm 0.055}$}} & 
        \multicolumn{1}{c}{\cellcolor{r5} \makecell{0.488$_{\pm 0.011}$ \\ 0.241$_{\pm 0.009}$}} & 
        \multicolumn{1}{c}{\cellcolor{r7} \makecell{0.431$_{\pm 0.044}$ \\ 0.200$_{\pm 0.022}$}} & 
        \multicolumn{1}{c}{\cellcolor{r2} \makecell{0.660$_{\pm 0.013}$ \\ 0.450$_{\pm 0.031}$}} & 
        \multicolumn{1}{c}{\cellcolor{r4} \makecell{0.622$_{\pm 0.069}$ \\ 0.423$_{\pm 0.039}$}} & 
        \multicolumn{1}{c}{\cellcolor{r3} \makecell{0.660$_{\pm 0.033}$ \\ 0.416$_{\pm 0.054}$}} & 
        \multicolumn{1}{c}{\cellcolor{r1} \makecell{0.663$_{\pm 0.038}$ \\ 0.456$_{\pm 0.048}$}} \\
        \midrule
        \makecell{SciBERT \\ \citep{beltagy2019scibert}} & 
        \multicolumn{1}{c}{\cellcolor{r7} \makecell{0.500$_{\pm 0.055}$ \\ 0.300$_{\pm 0.080}$}} & 
        \multicolumn{1}{c}{\cellcolor{r6} \makecell{0.502$_{\pm 0.030}$ \\ 0.248$_{\pm 0.015}$}} & 
        \multicolumn{1}{c}{\cellcolor{r5} \makecell{0.504$_{\pm 0.052}$ \\ 0.275$_{\pm 0.031}$}} & 
        \multicolumn{1}{c}{\cellcolor{r3} \makecell{0.680$_{\pm 0.066}$ \\ 0.458$_{\pm 0.060}$}} & 
        \multicolumn{1}{c}{\cellcolor{r4} \makecell{0.660$_{\pm 0.042}$ \\ 0.435$_{\pm 0.061}$}} & 
        \multicolumn{1}{c}{\cellcolor{r1} \makecell{0.686$_{\pm 0.039}$ \\ 0.460$_{\pm 0.042}$}} & 
        \multicolumn{1}{c}{\cellcolor{r2} \makecell{0.685$_{\pm 0.056}$ \\ 0.460$_{\pm 0.044}$}} \\
        \midrule
        \makecell{ScholarBERT \\ \citep{hong2022scholarbert}} & 
        \multicolumn{1}{c}{\cellcolor{r3} \makecell{0.472$_{\pm 0.137}$ \\ 0.234$_{\pm 0.094}$}} & 
        \multicolumn{1}{c}{\cellcolor{r6} \makecell{0.429$_{\pm 0.258}$ \\ 0.250$_{\pm 0.142}$}} & 
        \multicolumn{1}{c}{\cellcolor{r7} \makecell{0.367$_{\pm 0.075}$ \\ 0.165$_{\pm 0.044}$}} & 
        \multicolumn{1}{c}{\cellcolor{r5} \makecell{0.461$_{\pm 0.016}$ \\ 0.271$_{\pm 0.022}$}} & 
        \multicolumn{1}{c}{\cellcolor{r1} \makecell{0.513$_{\pm 0.041}$ \\ 0.295$_{\pm 0.055}$}} & 
        \multicolumn{1}{c}{\cellcolor{r4} \makecell{0.467$_{\pm 0.019}$ \\ 0.260$_{\pm 0.018}$}} & 
        \multicolumn{1}{c}{\cellcolor{r2} \makecell{0.468$_{\pm 0.028}$ \\ 0.276$_{\pm 0.024}$}} \\
        \midrule
        \makecell{BioBERT \\ \citep{shoyabiobert}} & 
        \multicolumn{1}{c}{\cellcolor{r6} \makecell{0.487$_{\pm 0.059}$ \\ 0.281$_{\pm 0.026}$}} & 
        \multicolumn{1}{c}{\cellcolor{r5} \makecell{0.488$_{\pm 0.032}$ \\ 0.238$_{\pm 0.017}$}} & 
        \multicolumn{1}{c}{\cellcolor{r7} \makecell{0.360$_{\pm 0.007}$ \\ 0.151$_{\pm 0.002}$}} & 
        \multicolumn{1}{c}{\cellcolor{r2} \makecell{0.663$_{\pm 0.044}$ \\ 0.442$_{\pm 0.079}$}} & 
        \multicolumn{1}{c}{\cellcolor{r4} \makecell{0.587$_{\pm 0.022}$ \\ 0.365$_{\pm 0.018}$}} & 
        \multicolumn{1}{c}{\cellcolor{r3} \makecell{0.632$_{\pm 0.040}$ \\ 0.404$_{\pm 0.046}$}} & 
        \multicolumn{1}{c}{\cellcolor{r1} \makecell{0.670$_{\pm 0.061}$ \\ 0.442$_{\pm 0.057}$}} \\
        \midrule
        \makecell{BERT \\ \citep{devlin2018bert}} & 
        \multicolumn{1}{c}{\cellcolor{r5} \makecell{0.498$_{\pm 0.051}$ \\ 0.266$_{\pm 0.044}$}} & 
        \multicolumn{1}{c}{\cellcolor{r6} \makecell{0.488$_{\pm 0.043}$ \\ 0.239$_{\pm 0.011}$}} & 
        \multicolumn{1}{c}{\cellcolor{r7} \makecell{0.394$_{\pm 0.009}$ \\ 0.166$_{\pm 0.008}$}} & 
        \multicolumn{1}{c}{\cellcolor{r1} \makecell{0.670$_{\pm 0.020}$ \\ 0.440$_{\pm 0.052}$}} & 
        \multicolumn{1}{c}{\cellcolor{r4} \makecell{0.601$_{\pm 0.046}$ \\ 0.382$_{\pm 0.039}$}} & 
        \multicolumn{1}{c}{\cellcolor{r3} \makecell{0.636$_{\pm 0.052}$ \\ 0.394$_{\pm 0.051}$}} & 
        \multicolumn{1}{c}{\cellcolor{r2} \makecell{0.658$_{\pm 0.030}$ \\ 0.439$_{\pm 0.021}$}} \\
        \midrule
        \makecell{Overall \\ (All Models) } & 
        \multicolumn{1}{c}{\cellcolor{r5} \makecell{0.493$_{\pm 0.064}$ \\ 0.288$_{\pm 0.063}$}} & 
        \multicolumn{1}{c}{\cellcolor{r6} \makecell{0.486$_{\pm 0.062}$ \\ 0.246$_{\pm 0.032}$}} & 
        \multicolumn{1}{c}{\cellcolor{r7} \makecell{0.439$_{\pm 0.003}$ \\ 0.212$_{\pm 0.022}$}} & 
        \multicolumn{1}{c}{\cellcolor{r2} \makecell{0.644$_{\pm 0.034}$ \\ 0.430$_{\pm 0.049}$}} & 
        \multicolumn{1}{c}{\cellcolor{r4} \makecell{0.622$_{\pm 0.035}$ \\ 0.402$_{\pm 0.049}$}} & 
        \multicolumn{1}{c}{\cellcolor{r3} \makecell{0.639$_{\pm 0.044}$ \\ 0.410$_{\pm 0.043}$}} & 
        \multicolumn{1}{c}{\cellcolor{r1} \makecell{0.688$_{\pm 0.046}$ \\ 0.435$_{\pm 0.039}$}} \\
        \bottomrule
    \end{tabular}
\end{adjustbox}
\end{spacing}
\vspace{-2mm}
\caption{Consolidated results among all MatSci-NLP tasks on different training settings for various BERT-based encoder models pretrained on different domain specific text data. For each model, the top line represents the micro-F1 score and the bottom line represents the macro-F1 score. We report the mean across 5 experiments with a confidence interval of two standard deviations. We highlight the performance of different schema according to heatmap ranging from \hlc[r1]{best} and \hlc[r7]{worst}. The concentration of red hues on right side indicates that the question-answering inspiring schema generally outperform conventional fine-tuning method. Our proposed \schema generally outperforms all other schemas across most enconder models.}
\label{table:schema-setting}
\end{table*}


Our analysis focuses on the questions outlined in \Cref{sec:intro}: 1) Studying the effectiveness of domain-specific language models as encoders, and 2) Analyzing the effect of different input schemas in resolving \benchmark tasks.
Concretely, we study the performance of the language models and language schema in a \textit{low resource} setting where we perform fine-tuning on different pretrained BERT models with limited data from the MatSci-NLP benchmark. This low-resource setting makes the learning problem harder given that the model has to generalize on little amount of data. Moreover, this setting approximates model training with very limited annotated data, which is commonly found in materials science as discussed in \Cref{sec:background}. In our experiments, we split the data in \benchmark into 1\% training subset and a 99\% testing subset for evaluation.
None of the evaluated encoder models were exposed to the fine-tuning data in advance of our experiments and therefore have to rely on the knowledge acquired during their respective pretraining processes.
We evaluate the results of our experiments using micro-F1 and macro-F1 scores of the language model predictions on the test split of the \benchmark that were not exposed during fine-tuning.


\subsection{How does in-domain pretraining of language models affect the downstream performance on MatSci-NLP tasks? ~(\ref{q1})}

Based on the results shown in \Cref{table:low-resource-setting}, we can gather the following insights:

\textit{First, domain-specific pretraining affects model performance.} We perform fine-tuning on various models pretrained on domain-specific corpora in a low-resource setting and observe that:
i) MatBert, which was pretrained on textual data from materials science journals, generally performs best for most tasks in the MatSci-NLP benchmark with SciBERT generally performing second best. 
The high performance of MatBERT suggests that materials science specific pretraining does help the language models acquire relevant materials science knowledge. Yet, the underperformance of MatSciBERT compared to MatBERT and SciBERT indicates that the curation of pretraining data does significantly affect performance.
ii) The importance of the pretraining corpus is further reinforced by the difference in performance between SciBERT and ScholarBERT, both of which were trained on corpora of general scientific text, but show vastly different results. In fact, ScholarBERT underperforms all other models, including the general language BERT, for all tasks except event argument extraction where ScholarBERT performs best compared to all other models. 
iii) The fact that most scientific BERT models outperform BERT pretrained on general language suggests that pretraining on high-quality scientific text is beneficial for resolving tasks involving materials science text and potentially scientific texts from other domains.
This notion of enhanced performance on \benchmark when pretraining on scientific text is further reinforced by the performance of BioBERT by \citet{shoyabiobert}. BioBERT outperforms BERT on most tasks even though it was trained on text from the biomedical domain that has minor overlap with the materials science domain. This strongly indicates that scientific language, regardless of the domain, has a significant distribution shift from general language that is used to pretrain common language models.

\textit{Second, imbalanced datasets in \benchmark skew performance metrics:} We can see from \Cref{table:low-resource-setting} that the micro-F1 scores are significantly higher than the macro-f1 across all tasks. This indicates that the datasets used in the MatSci-NLP are consistently imbalanced, including in the binary classification tasks, and thereby push the micro-F1 higher compared to the macro-F1 score. In the case of paragraph classification, for example, the number of positive examples is 492 compared with the total number of 1500 samples. As such, only models with a micro-F1 score above 0.66 and macro-F1 above 0.5 can be considered to have semantically meaningful understanding of the task. This is even more pronounced for sentence classification where only  $876/9466 \approx 10\%$ corresponds to one label. All models except ScholarBERT outperform a default guess of the dominant class for cases. While imbalanced datasets may approximate some real-world use cases of materials science text analysis, such as extracting specialized materials information, a highly imbalanced can be misguiding in evaluating model performance. 

To alleviate the potentially negative effects of imbalanced data, we suggest three simple yet effective methods: 1) Weighted loss functions: This involves weighting the loss function to give higher weights to minority classes. Focal loss~\cite{lin2017focal}, for example, is a loss function that dynamically modulates the loss based on the prediction confidence, with greater emphasis on more difficult examples. As such, Focal loss handles class imbalance well due to the additional attention given to hard examples of the minority classes. 2) Class-balanced samplers: Deep learning frameworks, such as Pytorch, have class-balanced batch samplers that can be used to oversample minority classes within each batch during training, which can help indirectly address class imbalance. 3) Model architecture tweaks: The model architecture and its hyper-parameters can be adjusted to place greater emphasis on minority classes. For example, one can apply separate prediction heads for minority classes or tweak L2 regularization and dropout to behave differently for minority and majority classes.


\subsection{How do in-context data schema and multitasking affect the learning efficiency in low-resource training settings? ~(\ref{q2})} \label{sec:results-q2}

To assess the efficacy of the proposed textual schemas shown in \Cref{fig:matsci-nlp-example}, we evaluate four different QA-inspired schemas: 1) \textit{No Explanations} - here the model receives only the task description; 2) \textit{Potential Choices} - here the model receives the class labels given by the task; 3) \textit{Examples} - here the model receives an example of a correct answer, 4) \textit{\schema}- here the model receives our proposed textual schema. We compare the schemas to three conventional fine-tuning methods: 1) \textit{Single Task} - the traditional method to solve each task separately using the language model and a classification head; 2) \textit{Single Task Prompt} - here we change the format of the task to the same QA-format as ``No Explanations'', but train each task separately; 3) \textit{MMOE} by \citet{ma2018modeling} uses multiple encoders to learn multiple hidden embeddings, which are then weighed by a task-specific gate unit and aggregated to the final hidden embedding using a weighted sum for each task. Next, a task-specific classification head outputs the label probability distribution for each task. 

Based on the results shown in \Cref{table:schema-setting}, we gather the following insights:

\textit{First, Text-to-Schema methods perform better for all language models.} Overall, the \schema method we proposed performs best across all tasks in the MatSci-NLP benchmark. The question-answering inspired schema (``No Explanations'', ``Potential Choices'', ``Examples'', ``\schema'') perform better than fine-tuning in a traditional single task setting, single task prompting, as well as fine-tuning using the MMOE multitask method. This holds across all models for all the tasks in \benchmark showing the efficacy of structured language modeling inspired by question-answering.

\textit{Second, schema design affects model performance.} The results show that both the pretrained model and the input format affect performance. This can be seen by the fact that while all scientific models outperform general language BERT using the \schema method, BERT outperforms some models, mainly ScholarBERT and BioBERT, in the other text-to-schema settings and the conventional training settings. Nevertheless, BERT underperforms the stronger models (MatBERT, SciBERT, MatSciBERT) across all schema settings for all tasks in MatSci-NLP, further emphasizing the importance of domain-specific model pretraining for materials science language understanding.

\section{Conclusion and Future Works}
We proposed MatSci-NLP, the first broad benchmark on materials science language understanding tasks constructed from publicly available data.
We further proposed text-to-schema multitask modeling to improve the model performance in low-resource settings.
Leveraging MatSci-NLP and text-to-schema modeling, we performed an in-depth analysis of the performance of various scientific language models and compare text-to-schema language modeling methods with other input schemas, guided by ~(\ref{q1}) addressing the pretrained models and ~(\ref{q2}) addressing the textual schema. 
Overall, we found that the choice of pretrained models matters significantly for downstream performance on MatSci-NLP tasks and that pretrained language models on scientific text of any kind often perform better than pretrained language models on general text. MatBERT generally performed best, highlighting the benefits of pretraining with high-quality domain-specific language data. 
With regards to the textual schema outlined in ~(\ref{q2}), we found that significant improvements can be made by improving textual schema showcasing the potential of fine-tuning using structured language modeling.

The proposed encoder-decoder architecture, as well as the proposed multitask schema, could also be useful for additional domains in NLP, including both scientific and non-scientific domains. The potential for open-domain transferability of our method is due to:
1) Our multitask training method and associated schemas do not depend on any domain-specific knowledge, allowing them to be easily transferred to other domains. 
2) The encoder of our proposed model architecture can be exchanged in a modular manner, which enables our model structure to be applied across multiple domains.
3) If the fine-tuning data is diverse across a wide range of domains, our method is likely to learn general language representations for open-domain multitask problems. 
Future work could build upon this paper by applying the model and proposed schema to different scientific domains where fine-tuning data might be sparse, such as biology, physics and chemistry. Moreover, future work can build upon the proposed schema by suggesting novel ways of modeling domain-specific or general language that lead to improvements in unified multi-task learning.


\section*{Limitations} \label{sec:limitations}
One of the primary limitations of NLP modeling in materials science, including this work, is the low quantity of available data as discussed in \Cref{sec:background}. This analysis is affected by this limitation as well given that our evaluations were performed in a low-data setting within a dataset that was already limited in size. We believe that future work can improve upon this study by applying larger datasets, both in the number of samples and in the scope of tasks, to similar problem settings. The small nature of the datasets applied in this study also presents the danger that some of the models may have memorized certain answers instead of achieving a broader understanding, which could be mitigated by enlarging the datasets and making the tasks more complex. 

Moreover, we did not study the generalization of NLP models beyond the materials science domain, including adjacent domains such as chemistry and physics. This targeted focus was intentional but imposes limitations on whether the proposed techniques and insights we gained from our analysis are transferable to other domains, including applying NLP models for scientific tasks outside of materials science.

Another limitation of our study is the fact that we focused on BERT-based models exclusively and did not study autoregressive models, including large language models with billions of parameters highlighted in the introduction. The primary reason for focusing on BERT-based models was the diversity of available models trained on different scientific text corpora. Large autoregressive models, on the other hand, are mostly trained on general text corpora with some notable exceptions, such as Galactica \citep{taylor2022galactica}. We believe that future work analyzing a greater diversity of language models, including large autoregressive models pretrained on different kinds of text, would significantly strengthen the understanding surrounding the ability of NLP models to perform text-based tasks in materials science. 

While the results presented in this study indicate that domain-specific pretraining can lead to noticeable advantages in downstream performance on text-based materials science tasks, we would like to highlight the associated risks and costs of pretraining a larger set of customized language models for different domains. The heavy financial and environmental costs associated with these pretraining procedures merit careful consideration of what conditions may warrant expensive pretraining and which ones may not. When possible, we encourage future researchers to build upon existing large models to mitigate the pretraining costs.

\section*{Broader Impacts and Ethics Statement}
Our MatSci-NLP benchmark can help promote the research on NLP for material science, an important and growing research field.
We expect that the experience we gained from the material science domain can be transferred to other domains, such as biology, health, and chemistry.
Our Text-to-Schema also helps with improving NLP tasks' performance in low-resource situations, which is a common challenge in many fields. 

Our research does not raise major ethical concerns.

\section*{Acknowlegments}
This work is supported by the Mila internal funding - Program P2-V1: Industry Sponsored Academic Labs (project number: 10379), the Canada CIFAR AI Chair Program, and the Canada NSERC Discovery Grant (RGPIN-2021-03115).

\bibliography{nlp4matsci}

\begin{thebibliography}{47}
\expandafter\ifx\csname natexlab\endcsname\relax\def\natexlab#1{#1}\fi

\bibitem[{Balikas et~al.(2015)Balikas, Krithara, Partalas, and
  Paliouras}]{balikas2015bioasq}
Georgios Balikas, Anastasia Krithara, Ioannis Partalas, and George Paliouras.
  2015.
\newblock Bioasq: A challenge on large-scale biomedical semantic indexing and
  question answering.
\newblock In \emph{International Workshop on Multimodal Retrieval in the
  Medical Domain}, pages 26--39. Springer.

\bibitem[{Beltagy et~al.(2019)Beltagy, Lo, and Cohan}]{beltagy2019scibert}
Iz~Beltagy, Kyle Lo, and Arman Cohan. 2019.
\newblock Scibert: A pretrained language model for scientific text.
\newblock \emph{arXiv preprint arXiv:1903.10676}.

\bibitem[{Brown et~al.(2020)Brown, Mann, Ryder, Subbiah, Kaplan, Dhariwal,
  Neelakantan, Shyam, Sastry, Askell et~al.}]{brown2020language}
Tom Brown, Benjamin Mann, Nick Ryder, Melanie Subbiah, Jared~D Kaplan, Prafulla
  Dhariwal, Arvind Neelakantan, Pranav Shyam, Girish Sastry, Amanda Askell,
  et~al. 2020.
\newblock Language models are few-shot learners.
\newblock \emph{Advances in neural information processing systems},
  33:1877--1901.

\bibitem[{Choudhary et~al.(2022)Choudhary, DeCost, Chen, Jain, Tavazza, Cohn,
  Park, Choudhary, Agrawal, Billinge et~al.}]{choudhary2022recent}
Kamal Choudhary, Brian DeCost, Chi Chen, Anubhav Jain, Francesca Tavazza, Ryan
  Cohn, Cheol~Woo Park, Alok Choudhary, Ankit Agrawal, Simon~JL Billinge,
  et~al. 2022.
\newblock Recent advances and applications of deep learning methods in
  materials science.
\newblock \emph{npj Computational Materials}, 8(1):1--26.

\bibitem[{Dasigi et~al.(2021)Dasigi, Lo, Beltagy, Cohan, Smith, and
  Gardner}]{dasigi2021dataset}
Pradeep Dasigi, Kyle Lo, Iz~Beltagy, Arman Cohan, Noah~A Smith, and Matt
  Gardner. 2021.
\newblock A dataset of information-seeking questions and answers anchored in
  research papers.
\newblock \emph{arXiv preprint arXiv:2105.03011}.

\bibitem[{Devlin et~al.(2018)Devlin, Chang, Lee, and
  Toutanova}]{devlin2018bert}
Jacob Devlin, Ming-Wei Chang, Kenton Lee, and Kristina Toutanova. 2018.
\newblock Bert: Pre-training of deep bidirectional transformers for language
  understanding.
\newblock \emph{arXiv preprint arXiv:1810.04805}.

\bibitem[{Friedrich et~al.(2020)Friedrich, Adel, Tomazic, Hingerl, Benteau,
  Maruscyk, and Lange}]{friedrich2020sofc}
Annemarie Friedrich, Heike Adel, Federico Tomazic, Johannes Hingerl, Renou
  Benteau, Anika Maruscyk, and Lukas Lange. 2020.
\newblock The sofc-exp corpus and neural approaches to information extraction
  in the materials science domain.
\newblock \emph{arXiv preprint arXiv:2006.03039}.

\bibitem[{Georgescu et~al.(2021)Georgescu, Ren, Toland, Zhang, Miller, Apley,
  Olivetti, Wagner, and Rondinelli}]{georgescu2021database}
Alexandru~B Georgescu, Peiwen Ren, Aubrey~R Toland, Shengtong Zhang, Kyle~D
  Miller, Daniel~W Apley, Elsa~A Olivetti, Nicholas Wagner, and James~M
  Rondinelli. 2021.
\newblock Database, features, and machine learning model to identify thermally
  driven metal--insulator transition compounds.
\newblock \emph{Chemistry of Materials}, 33(14):5591--5605.

\bibitem[{Gu et~al.(2021)Gu, Tinn, Cheng, Lucas, Usuyama, Liu, Naumann, Gao,
  and Poon}]{gu2021domain}
Yu~Gu, Robert Tinn, Hao Cheng, Michael Lucas, Naoto Usuyama, Xiaodong Liu,
  Tristan Naumann, Jianfeng Gao, and Hoifung Poon. 2021.
\newblock Domain-specific language model pretraining for biomedical natural
  language processing.
\newblock \emph{ACM Transactions on Computing for Healthcare (HEALTH)},
  3(1):1--23.

\bibitem[{Gupta et~al.(2022)Gupta, Zaki, Krishnan et~al.}]{gupta2022matscibert}
Tanishq Gupta, Mohd Zaki, NM~Krishnan, et~al. 2022.
\newblock Matscibert: A materials domain language model for text mining and
  information extraction.
\newblock \emph{npj Computational Materials}, 8(1):1--11.

\bibitem[{Hakala and Pyysalo(2019)}]{hakala2019biomedical}
Kai Hakala and Sampo Pyysalo. 2019.
\newblock Biomedical named entity recognition with multilingual bert.
\newblock In \emph{Proceedings of the 5th workshop on BioNLP open shared
  tasks}, pages 56--61.

\bibitem[{Hong et~al.(2022)Hong, Ajith, Pauloski, Duede, Malamud, Magoulas,
  Chard, and Foster}]{hong2022scholarbert}
Zhi Hong, Aswathy Ajith, Gregory Pauloski, Eamon Duede, Carl Malamud, Roger
  Magoulas, Kyle Chard, and Ian Foster. 2022.
\newblock Scholarbert: Bigger is not always better.
\newblock \emph{arXiv preprint arXiv:2205.11342}.

\bibitem[{Huang and Cole(2022)}]{huang2022batterybert}
Shu Huang and Jacqueline~M Cole. 2022.
\newblock Batterybert: A pretrained language model for battery database
  enhancement.
\newblock \emph{Journal of Chemical Information and Modeling}.

\bibitem[{Jensen et~al.(2021)Jensen, Kwon, Schwalbe-Koda, Paris,
  G{\'o}mez-Bombarelli, Rom{\'a}n-Leshkov, Corma, Moliner, and
  Olivetti}]{jensen2021discovering}
Zach Jensen, Soonhyoung Kwon, Daniel Schwalbe-Koda, Cecilia Paris, Rafael
  G{\'o}mez-Bombarelli, Yuriy Rom{\'a}n-Leshkov, Avelino Corma, Manuel Moliner,
  and Elsa~A Olivetti. 2021.
\newblock Discovering relationships between osdas and zeolites through data
  mining and generative neural networks.
\newblock \emph{ACS central science}, 7(5):858--867.

\bibitem[{Jin et~al.(2019)Jin, Dhingra, Liu, Cohen, and Lu}]{jin2019pubmedqa}
Qiao Jin, Bhuwan Dhingra, Zhengping Liu, William~W Cohen, and Xinghua Lu. 2019.
\newblock Pubmedqa: A dataset for biomedical research question answering.
\newblock \emph{arXiv preprint arXiv:1909.06146}.

\bibitem[{Karpovich et~al.(2021)Karpovich, Jensen, Venugopal, and
  Olivetti}]{karpovich2021inorganic}
Christopher Karpovich, Zach Jensen, Vineeth Venugopal, and Elsa Olivetti. 2021.
\newblock Inorganic synthesis reaction condition prediction with generative
  machine learning.
\newblock \emph{arXiv preprint arXiv:2112.09612}.

\bibitem[{Kim et~al.(2020)Kim, Jensen, van Grootel, Huang, Staib, Mysore,
  Chang, Strubell, McCallum, Jegelka et~al.}]{kim2020inorganic}
Edward Kim, Zach Jensen, Alexander van Grootel, Kevin Huang, Matthew Staib,
  Sheshera Mysore, Haw-Shiuan Chang, Emma Strubell, Andrew McCallum, Stefanie
  Jegelka, et~al. 2020.
\newblock Inorganic materials synthesis planning with literature-trained neural
  networks.
\newblock \emph{Journal of chemical information and modeling},
  60(3):1194--1201.

\bibitem[{Kingma and Ba(2014)}]{kingma2014adam}
Diederik~P Kingma and Jimmy Ba. 2014.
\newblock Adam: A method for stochastic optimization.
\newblock \emph{arXiv preprint arXiv:1412.6980}.

\bibitem[{Kononova et~al.(2021)Kononova, He, Huo, Trewartha, Olivetti, and
  Ceder}]{kononova2021opportunities}
Olga Kononova, Tanjin He, Haoyan Huo, Amalie Trewartha, Elsa~A Olivetti, and
  Gerbrand Ceder. 2021.
\newblock Opportunities and challenges of text mining in materials research.
\newblock \emph{Iscience}, 24(3):102155.

\bibitem[{Kuniyoshi et~al.(2020)Kuniyoshi, Makino, Ozawa, and
  Miwa}]{kuniyoshi2020annotating}
Fusataka Kuniyoshi, Kohei Makino, Jun Ozawa, and Makoto Miwa. 2020.
\newblock Annotating and extracting synthesis process of all-solid-state
  batteries from scientific literature.
\newblock \emph{arXiv preprint arXiv:2002.07339}.

\bibitem[{Lee et~al.(2020)Lee, Yoon, Kim, Kim, Kim, So, and
  Kang}]{lee2020biobert}
Jinhyuk Lee, Wonjin Yoon, Sungdong Kim, Donghyeon Kim, Sunkyu Kim, Chan~Ho So,
  and Jaewoo Kang. 2020.
\newblock Biobert: a pre-trained biomedical language representation model for
  biomedical text mining.
\newblock \emph{Bioinformatics}, 36(4):1234--1240.

\bibitem[{Lin et~al.(2017)Lin, Goyal, Girshick, He, and
  Doll{\'a}r}]{lin2017focal}
Tsung-Yi Lin, Priya Goyal, Ross Girshick, Kaiming He, and Piotr Doll{\'a}r.
  2017.
\newblock Focal loss for dense object detection.
\newblock In \emph{Proceedings of the IEEE international conference on computer
  vision}, pages 2980--2988.

\bibitem[{Lu et~al.(2021)Lu, Lin, Xu, Han, Tang, Li, Sun, Liao, and
  Chen}]{lu2021text2event}
Yaojie Lu, Hongyu Lin, Jin Xu, Xianpei Han, Jialong Tang, Annan Li, Le~Sun,
  Meng Liao, and Shaoyi Chen. 2021.
\newblock Text2event: Controllable sequence-to-structure generation for
  end-to-end event extraction.
\newblock \emph{arXiv preprint arXiv:2106.09232}.

\bibitem[{Luong et~al.(2015)Luong, Le, Sutskever, Vinyals, and
  Kaiser}]{luong2015multi}
Minh-Thang Luong, Quoc~V Le, Ilya Sutskever, Oriol Vinyals, and Lukasz Kaiser.
  2015.
\newblock Multi-task sequence to sequence learning.
\newblock \emph{arXiv preprint arXiv:1511.06114}.

\bibitem[{Ma et~al.(2018)Ma, Zhao, Yi, Chen, Hong, and Chi}]{ma2018modeling}
Jiaqi Ma, Zhe Zhao, Xinyang Yi, Jilin Chen, Lichan Hong, and Ed~H Chi. 2018.
\newblock Modeling task relationships in multi-task learning with multi-gate
  mixture-of-experts.
\newblock In \emph{Proceedings of the 24th ACM SIGKDD international conference
  on knowledge discovery \& data mining}, pages 1930--1939.

\bibitem[{Mahbub et~al.(2020)Mahbub, Huang, Jensen, Hood, Rupp, and
  Olivetti}]{mahbub2020text}
Rubayyat Mahbub, Kevin Huang, Zach Jensen, Zachary~D Hood, Jennifer~LM Rupp,
  and Elsa~A Olivetti. 2020.
\newblock Text mining for processing conditions of solid-state battery
  electrolytes.
\newblock \emph{Electrochemistry Communications}, 121:106860.

\bibitem[{MatSciRE(2022)}]{matscire2022github}
MatSciRE. 2022.
\newblock \href
  {https://github.com/MatSciRE/Material_Science_Relation_Extraction} {Material
  science relation extraction (matscire)}.

\bibitem[{Miret et~al.()Miret, Skreta, Sanchez-Lengelin, Ong, Morgan-Chan, and
  Aspuru-Guzik}]{ai4mat}
Santiago Miret, Marta Skreta, Benjamin Sanchez-Lengelin, Shyue~Ping Ong, Zamyla
  Morgan-Chan, and Alan Aspuru-Guzik.
\newblock \href {https://sites.google.com/view/ai4mat} {Ai4mat - neurips 2022}.

\bibitem[{Mysore et~al.(2019)Mysore, Jensen, Kim, Huang, Chang, Strubell,
  Flanigan, McCallum, and Olivetti}]{mysore2019materials}
Sheshera Mysore, Zach Jensen, Edward Kim, Kevin Huang, Haw-Shiuan Chang, Emma
  Strubell, Jeffrey Flanigan, Andrew McCallum, and Elsa Olivetti. 2019.
\newblock The materials science procedural text corpus: Annotating materials
  synthesis procedures with shallow semantic structures.
\newblock \emph{arXiv preprint arXiv:1905.06939}.

\bibitem[{Olivetti et~al.(2020)Olivetti, Cole, Kim, Kononova, Ceder, Han, and
  Hiszpanski}]{olivetti2020data}
Elsa~A Olivetti, Jacqueline~M Cole, Edward Kim, Olga Kononova, Gerbrand Ceder,
  Thomas Yong-Jin Han, and Anna~M Hiszpanski. 2020.
\newblock Data-driven materials research enabled by natural language processing
  and information extraction.
\newblock \emph{Applied Physics Reviews}, 7(4):041317.

\bibitem[{Peng et~al.(2019)Peng, Yan, and Lu}]{peng2019transfer}
Yifan Peng, Shankai Yan, and Zhiyong Lu. 2019.
\newblock Transfer learning in biomedical natural language processing: An
  evaluation of bert and elmo on ten benchmarking datasets.
\newblock In \emph{Proceedings of the 2019 Workshop on Biomedical Natural
  Language Processing (BioNLP 2019)}.

\bibitem[{Phan et~al.(2021)Phan, Anibal, Tran, Chanana, Bahadroglu, Peltekian,
  and Altan-Bonnet}]{phan2021scifive}
Long~N Phan, James~T Anibal, Hieu Tran, Shaurya Chanana, Erol Bahadroglu, Alec
  Peltekian, and Gr{\'e}goire Altan-Bonnet. 2021.
\newblock Scifive: a text-to-text transformer model for biomedical literature.
\newblock \emph{arXiv preprint arXiv:2106.03598}.

\bibitem[{Pilania(2021)}]{pilania2021machine}
Ghanshyam Pilania. 2021.
\newblock Machine learning in materials science: From explainable predictions
  to autonomous design.
\newblock \emph{Computational Materials Science}, 193:110360.

\bibitem[{Qu et~al.(2019)Qu, Yang, Qiu, Croft, Zhang, and Iyyer}]{qu2019bert}
Chen Qu, Liu Yang, Minghui Qiu, W~Bruce Croft, Yongfeng Zhang, and Mohit Iyyer.
  2019.
\newblock Bert with history answer embedding for conversational question
  answering.
\newblock In \emph{Proceedings of the 42nd international ACM SIGIR conference
  on research and development in information retrieval}, pages 1133--1136.

\bibitem[{Raffel et~al.(2020)Raffel, Shazeer, Roberts, Lee, Narang, Matena,
  Zhou, Li, Liu et~al.}]{raffel2020exploring}
Colin Raffel, Noam Shazeer, Adam Roberts, Katherine Lee, Sharan Narang, Michael
  Matena, Yanqi Zhou, Wei Li, Peter~J Liu, et~al. 2020.
\newblock Exploring the limits of transfer learning with a unified text-to-text
  transformer.
\newblock \emph{J. Mach. Learn. Res.}, 21(140):1--67.

\bibitem[{Scao et~al.(2022)Scao, Fan, Akiki, Pavlick, Ili{\'c}, Hesslow,
  Castagn{\'e}, Luccioni, Yvon, Gall{\'e} et~al.}]{scao2022bloom}
Teven~Le Scao, Angela Fan, Christopher Akiki, Ellie Pavlick, Suzana Ili{\'c},
  Daniel Hesslow, Roman Castagn{\'e}, Alexandra~Sasha Luccioni, Fran{\c{c}}ois
  Yvon, Matthias Gall{\'e}, et~al. 2022.
\newblock Bloom: A 176b-parameter open-access multilingual language model.
\newblock \emph{arXiv preprint arXiv:2211.05100}.

\bibitem[{Shin et~al.(2020)Shin, Zhang, Bakhturina, Puri, Patwary, Shoeybi, and
  Mani}]{shin2020biomegatron}
Hoo-Chang Shin, Yang Zhang, Evelina Bakhturina, Raul Puri, Mostofa Patwary,
  Mohammad Shoeybi, and Raghav Mani. 2020.
\newblock Biomegatron: Larger biomedical domain language model.
\newblock \emph{arXiv preprint arXiv:2010.06060}.

\bibitem[{Taylor et~al.(2022)Taylor, Kardas, Cucurull, Scialom, Hartshorn,
  Saravia, Poulton, Kerkez, and Stojnic}]{taylor2022galactica}
Ross Taylor, Marcin Kardas, Guillem Cucurull, Thomas Scialom, Anthony
  Hartshorn, Elvis Saravia, Andrew Poulton, Viktor Kerkez, and Robert Stojnic.
  2022.
\newblock Galactica: A large language model for science.
\newblock \emph{arXiv preprint arXiv:2211.09085}.

\bibitem[{Van~Nguyen et~al.(2022)Van~Nguyen, Min, Dernoncourt, and
  Nguyen}]{van2022joint}
Minh Van~Nguyen, Bonan Min, Franck Dernoncourt, and Thien Nguyen. 2022.
\newblock Joint extraction of entities, relations, and events via modeling
  inter-instance and inter-label dependencies.
\newblock In \emph{Proceedings of the 2022 Conference of the North American
  Chapter of the Association for Computational Linguistics: Human Language
  Technologies}, pages 4363--4374.

\bibitem[{Venugopal et~al.(2021)Venugopal, Sahoo, Zaki, Agarwal, Gosvami, and
  Krishnan}]{venugopal2021looking}
Vineeth Venugopal, Sourav Sahoo, Mohd Zaki, Manish Agarwal, Nitya~Nand Gosvami,
  and NM~Anoop Krishnan. 2021.
\newblock Looking through glass: Knowledge discovery from materials science
  literature using natural language processing.
\newblock \emph{Patterns}, 2(7):100290.

\bibitem[{Wada et~al.(2020)Wada, Takeda, Manabe, Konishi, Kamohara, and
  Matsumura}]{shoyabiobert}
Shoya Wada, Toshihiro Takeda, Shiro Manabe, Shozo Konishi, Jun Kamohara, and
  Yasushi Matsumura. 2020.
\newblock \href {http://arxiv.org/abs/arXiv:2005.07202} {A pre-training
  technique to localize medical bert and enhance biobert}.

\bibitem[{Walker et~al.(2021)Walker, Trewartha, Huo, Lee, Cruse, Dagdelen,
  Dunn, Persson, Ceder, and Jain}]{walker2021impact}
Nicholas Walker, Amalie Trewartha, Haoyan Huo, Sanghoon Lee, Kevin Cruse, John
  Dagdelen, Alexander Dunn, Kristin Persson, Gerbrand Ceder, and Anubhav Jain.
  2021.
\newblock The impact of domain-specific pre-training on named entity
  recognition tasks in materials science.
\newblock \emph{Available at SSRN 3950755}.

\bibitem[{Wang et~al.(2022{\natexlab{a}})Wang, Cruse, Fei, Chia, Zeng, Huo, He,
  Deng, Kononova, and Ceder}]{wang2022ulsa}
Zheren Wang, Kevin Cruse, Yuxing Fei, Ann Chia, Yan Zeng, Haoyan Huo, Tanjin
  He, Bowen Deng, Olga Kononova, and Gerbrand Ceder. 2022{\natexlab{a}}.
\newblock Ulsa: Unified language of synthesis actions for the representation of
  inorganic synthesis protocols.
\newblock \emph{Digital Discovery}.

\bibitem[{Wang et~al.(2022{\natexlab{b}})Wang, Kononova, Cruse, He, Huo, Fei,
  Zeng, Sun, Cai, Sun et~al.}]{wang2022dataset}
Zheren Wang, Olga Kononova, Kevin Cruse, Tanjin He, Haoyan Huo, Yuxing Fei, Yan
  Zeng, Yingzhi Sun, Zijian Cai, Wenhao Sun, et~al. 2022{\natexlab{b}}.
\newblock Dataset of solution-based inorganic materials synthesis procedures
  extracted from the scientific literature.
\newblock \emph{Scientific Data}, 9(1):1--11.

\bibitem[{Weston et~al.(2019)Weston, Tshitoyan, Dagdelen, Kononova, Trewartha,
  Persson, Ceder, and Jain}]{weston2019named}
Leigh Weston, Vahe Tshitoyan, John Dagdelen, Olga Kononova, Amalie Trewartha,
  Kristin~A Persson, Gerbrand Ceder, and Anubhav Jain. 2019.
\newblock Named entity recognition and normalization applied to large-scale
  information extraction from the materials science literature.
\newblock \emph{Journal of chemical information and modeling},
  59(9):3692--3702.

\bibitem[{Wu and He(2019)}]{wu2019enriching}
Shanchan Wu and Yifan He. 2019.
\newblock Enriching pre-trained language model with entity information for
  relation classification.
\newblock In \emph{Proceedings of the 28th ACM international conference on
  information and knowledge management}, pages 2361--2364.

\bibitem[{Yamaguchi et~al.(2020)Yamaguchi, Asahi, and Sasaki}]{yamaguchi2020sc}
Kyosuke Yamaguchi, Ryoji Asahi, and Yutaka Sasaki. 2020.
\newblock Sc-comics: a superconductivity corpus for materials informatics.
\newblock In \emph{Proceedings of The 12th Language Resources and Evaluation
  Conference}, pages 6753--6760.

\end{thebibliography}
\bibliographystyle{acl_natbib}

\newpage
\appendix


\section*{Appendix}

\section{Experimental Details}
We performed fine-tuning experiments using a single GPU with a learning rate was 2e-5, the hidden size of the encoders being 768, except ScholarBERT which is 1024, using the Adam ~\cite{kingma2014adam} optimizer for a max number of 20 training epochs with early stopping. All models are implemented with Python and PyTorch, and repeated five times to report the average performance. The full set of hyperparameters will be provided in our code release upon publication. 

\section{Additional Text-to-Schema Experiments}
To arrive at our data presented in \Cref{table:schema-setting}, we conducted experiments for all the language models across all tasks in MatSci-NLP. The results for seven tasks in MatSci-NLP are shown in subsequent tables:
\begin{itemize}
    \item Named Entity Recognition in \Cref{table:schema-ner}.
    \item Relation Classification in \Cref{table:schema-re}.
    \item Event Argument Extraction in \Cref{table:schema-ee}.
    \item Paragraph Classification in \Cref{table:schema-pc}.
    \item Synthesis Action Retrieval in \Cref{table:schema-sar}.
    \item Sentence Classification in \Cref{table:schema-sc}.
    \item Slot Filling in \Cref{table:schema-sf}.
\end{itemize}
The experimental results summarized in the aforementioned tables reinforce the conclusions in our analysis of ~(\ref{q2}) in \Cref{sec:results-q2} with the text-to-schema based fine-tuning method generally outperforming the conventional single and multitask methods across all tasks and all language models.

\begin{table*}[h]
\begin{spacing}{1.5}
\centering
\vspace{-2.5mm}
\begin{adjustbox}{max width=1\linewidth}
    \begin{tabular}{c|c|c|c|c|c|c|c}
        \toprule
        \bf{NLP Model} & 
        \multicolumn{1}{c}{\makecell{\bf{Single Task}}} & 
        \multicolumn{1}{c}{\makecell{\bf{Single Task Prompt}}} &
        \multicolumn{1}{c}{\makecell{\bf{MMOE}}} & 
        \multicolumn{1}{c}{\makecell{\bf{No Explanations}}} & 
        \multicolumn{1}{c}{\makecell{\bf{Potential} \\ \bf{Choices}}} & 
        \multicolumn{1}{c}{\makecell{\bf{Examples}}} & 
        \multicolumn{1}{c}{\makecell{\bf{Text2Schema}}} \\
        \midrule
        \makecell{MatSciBERT \\ \citep{gupta2022matscibert}} & 
        \multicolumn{1}{c}{\makecell{0.690$_{\pm 0.018}$ \\ 0.403$_{\pm 0.029}$}} & 
        \multicolumn{1}{c}{\makecell{0.707$_{\pm 0.089}$ \\ 0.445$_{\pm 0.071}$}} & 
        \multicolumn{1}{c}{\makecell{0.451$_{\pm 0.114}$ \\ 0.188$_{\pm 0.065}$}} & 
        \multicolumn{1}{c}{\makecell{0.655$_{\pm 0.066}$ \\ 0.410$_{\pm 0.051}$}} & 
        \multicolumn{1}{c}{\makecell{0.732$_{\pm 0.048}$ \\ 0.480$_{\pm 0.087}$}} & 
        \multicolumn{1}{c}{ \cellcolor{r1} \makecell{0.753$_{\pm 0.060}$ \\ 0.505$_{\pm 0.066}$}} & 
        \multicolumn{1}{c}{\makecell{0.707$_{\pm 0.076}$ \\ 0.470$_{\pm 0.092}$}} \\
        \midrule
        \makecell{MatBERT \\ \citep{walker2021impact}} &  
        \multicolumn{1}{c}{\makecell{0.705$_{\pm 0.011}$ \\ 0.469$_{\pm 0.037}$}} & 
        \multicolumn{1}{c}{\makecell{0.796$_{\pm 0.029}$ \\ 0.558$_{\pm 0.044}$}} & 
        \multicolumn{1}{c}{\makecell{0.691$_{\pm 0.060}$ \\ 0.400$_{\pm 0.070}$}} & 
        \multicolumn{1}{c}{\cellcolor{r1} \makecell{0.805$_{\pm 0.018}$ \\ 0.574$_{\pm 0.061}$}} & 
        \multicolumn{1}{c}{\makecell{0.756$_{\pm 0.071}$ \\ 0.524$_{\pm 0.088}$}} & 
        \multicolumn{1}{c}{\makecell{0.778$_{\pm 0.015}$ \\ 0.547$_{\pm 0.039}$}} & 
        \multicolumn{1}{c}{\makecell{0.798$_{\pm 0.031}$ \\ 0.569$_{\pm 0.055}$}} \\
        \midrule
        \makecell{BatteryBERT \\ \citep{huang2022batterybert}} &  
        \multicolumn{1}{c}{\makecell{0.690$_{\pm 0.014}$ \\ 0.464$_{\pm 0.018}$}} & 
        \multicolumn{1}{c}{\makecell{0.673$_{\pm 0.029}$ \\ 0.407$_{\pm 0.045}$}} & 
        \multicolumn{1}{c}{\makecell{0.439$_{\pm 0.185}$ \\ 0.168$_{\pm 0.110}$}} & 
        \multicolumn{1}{c}{\makecell{0.733$_{\pm 0.026}$ \\ 0.483$_{\pm 0.049}$}} & 
        \multicolumn{1}{c}{\makecell{0.607$_{\pm 0.169}$ \\ 0.369$_{\pm 0.140}$}} & 
        \multicolumn{1}{c}{\cellcolor{r1} \makecell{0.743$_{\pm 0.015}$ \\ 0.497$_{\pm 0.015}$}} & 
        \multicolumn{1}{c}{\makecell{0.722$_{\pm 0.045}$ \\ 0.470$_{\pm 0.043}$}} \\
        \midrule
        \makecell{SciBERT \\ \citep{beltagy2019scibert}} & 
        \multicolumn{1}{c}{\makecell{0.686$_{\pm 0.015}$ \\ 0.464$_{\pm 0.035}$}} & 
        \multicolumn{1}{c}{\cellcolor{r1} \makecell{0.754$_{\pm 0.029}$ \\ 0.493$_{\pm 0.063}$}} & 
        \multicolumn{1}{c}{\makecell{0.598$_{\pm 0.027}$ \\ 0.298$_{\pm 0.048}$}} & 
        \multicolumn{1}{c}{\makecell{0.708$_{\pm 0.115}$ \\ 0.465$_{\pm 0.115}$}} & 
        \multicolumn{1}{c}{\makecell{0.724$_{\pm 0.045}$ \\ 0.471$_{\pm 0.069}$}} & 
        \multicolumn{1}{c}{\cellcolor{r1} \makecell{0.754$_{\pm 0.054}$ \\ 0.509$_{\pm 0.064}$}} & 
        \multicolumn{1}{c}{\makecell{0.734$_{\pm 0.079}$ \\ 0.497$_{\pm 0.091}$}} \\
        \midrule
        \makecell{ScholarBERT \\ \citep{hong2022scholarbert}} & 
        \multicolumn{1}{c}{\makecell{0.206$_{\pm 0.350}$ \\ 0.069$_{\pm 0.131}$}} & 
        \multicolumn{1}{c}{\makecell{0.179$_{\pm 0.088}$ \\ 0.108$_{\pm 0.057}$}} & 
        \multicolumn{1}{c}{\makecell{0.109$_{\pm 0.142}$ \\ 0.018$_{\pm 0.033}$}} & 
        \multicolumn{1}{c}{\makecell{0.134$_{\pm 0.036}$ \\ 0.071$_{\pm 0.023}$}} & 
        \multicolumn{1}{c}{\cellcolor{r1} \makecell{0.263$_{\pm 0.109}$ \\ 0.122$_{\pm 0.073}$}} & 
        \multicolumn{1}{c}{\makecell{0.168$_{\pm 0.044}$ \\ 0.098$_{\pm 0.045}$}} & 
        \multicolumn{1}{c}{\makecell{0.168$_{\pm 0.067}$ \\ 0.101$_{\pm 0.034}$}} \\
        \midrule
        \makecell{BioBERT \\ \citep{shoyabiobert}} & 
        \multicolumn{1}{c}{\makecell{0.665$_{\pm 0.018}$ \\ 0.403$_{\pm 0.030}$}} & 
        \multicolumn{1}{c}{\makecell{0.708$_{\pm 0.119}$ \\ 0.431$_{\pm 0.115}$}} & 
        \multicolumn{1}{c}{\makecell{0.204$_{\pm 0.114}$ \\ 0.019$_{\pm 0.000}$}} & 
        \multicolumn{1}{c}{\makecell{0.723$_{\pm 0.075}$ \\ 0.474$_{\pm 0.071}$}} & 
        \multicolumn{1}{c}{\makecell{0.455$_{\pm 0.114}$ \\ 0.188$_{\pm 0.065}$}} & 
        \multicolumn{1}{c}{\cellcolor{r1} \makecell{0.725$_{\pm 0.024}$ \\ 0.452$_{\pm 0.044}$}} & 
        \multicolumn{1}{c}{\makecell{0.715$_{\pm 0.031}$ \\ 0.459$_{\pm 0.055}$}} \\
        \midrule
        \makecell{BERT \\ \citep{devlin2018bert}} & 
        \multicolumn{1}{c}{\makecell{0.606$_{\pm 0.009}$ \\ 0.304$_{\pm 0.024}$}} & 
        \multicolumn{1}{c}{\makecell{0.636$_{\pm 0.034}$ \\ 0.382$_{\pm 0.041}$}} & 
        \multicolumn{1}{c}{\makecell{0.235$_{\pm 0.069}$ \\ 0.055$_{\pm 0.040}$}} & 
        \multicolumn{1}{c}{\makecell{0.670$_{\pm 0.056}$ \\ 0.441$_{\pm 0.060}$}} & 
        \multicolumn{1}{c}{\makecell{0.455$_{\pm 0.138}$ \\ 0.267$_{\pm 0.089}$}} & 
        \multicolumn{1}{c}{\cellcolor{r1} \makecell{0.664$_{\pm 0.047}$ \\ 0.418$_{\pm 0.046}$}} & 
        \multicolumn{1}{c}{\makecell{0.657$_{\pm 0.079}$ \\ 0.416$_{\pm 0.058}$}} \\
        \bottomrule
    \end{tabular}
\end{adjustbox}
\end{spacing}
\vspace{-2mm}
\caption{Results of \textbf{named entity recognition} task among seven tasks on different schema settings for various BERT models pre-trained on different domain specific text data. For each model, the top line represents the micro-F1 score and the bottom line represents the macro-F1 score. We report the mean across 5 experiments with a confidence interval of two standard deviations. We highlight the \hlc[r1]{best} performing method. }
\label{table:schema-ner}
\end{table*}

\begin{table*}[h]
\begin{spacing}{1.5}
\centering
\vspace{-2.5mm}
\begin{adjustbox}{max width=1\linewidth}
    \begin{tabular}{c|c|c|c|c|c|c|c}
        \toprule
        \bf{NLP Model} & 
        \multicolumn{1}{c}{\makecell{\bf{Single Task}}} & 
        \multicolumn{1}{c}{\makecell{\bf{Single Task Prompt}}} &
        \multicolumn{1}{c}{\makecell{\bf{MMOE}}} & 
        \multicolumn{1}{c}{\makecell{\bf{No Explanations}}} & 
        \multicolumn{1}{c}{\makecell{\bf{Potential} \\ \bf{Choices}}} & 
        \multicolumn{1}{c}{\makecell{\bf{Examples}}} & 
        \multicolumn{1}{c}{\makecell{\bf{Text2Schema}}} \\
        \midrule
        \makecell{MatSciBERT \\ \citep{gupta2022matscibert}} & 
        \multicolumn{1}{c}{\makecell{0.671$_{\pm 0.083}$ \\ 0.439$_{\pm 0.137}$}} &
        \multicolumn{1}{c}{\makecell{0.545$_{\pm 0.102}$ \\ 0.219$_{\pm 0.035}$}} & 
        \multicolumn{1}{c}{\makecell{0.490$_{\pm 0.139}$ \\ 0.218$_{\pm 0.073}$}} & 
        \multicolumn{1}{c}{\makecell{0.747$_{\pm 0.128}$ \\ 0.461$_{\pm 0.190}$}} & 
        \multicolumn{1}{c}{\makecell{0.800$_{\pm 0.058}$ \\ 0.482$_{\pm 0.064}$}} & 
        \multicolumn{1}{c}{\cellcolor{r1} \makecell{0.818$_{\pm 0.137}$ \\ 0.530$_{\pm 0.203}$}} & 
        \multicolumn{1}{c}{\makecell{0.791$_{\pm 0.046}$ \\ 0.507$_{\pm 0.073}$}} \\
        \midrule
        \makecell{MatBERT \\ \citep{walker2021impact}} &  
        \multicolumn{1}{c}{\makecell{0.714$_{\pm 0.023}$ \\ 0.487$_{\pm 0.075}$}} &
        \multicolumn{1}{c}{\makecell{0.644$_{\pm 0.050}$ \\ 0.310$_{\pm 0.078}$}} & 
        \multicolumn{1}{c}{\makecell{0.591$_{\pm 0.267}$ \\ 0.297$_{\pm 0.143}$}} & 
        \multicolumn{1}{c}{\makecell{0.871$_{\pm 0.020}$ \\ 0.623$_{\pm 0.035}$}} & 
        \multicolumn{1}{c}{\makecell{0.804$_{\pm 0.071}$ \\ 0.513$_{\pm 0.138}$}} & 
        \multicolumn{1}{c}{\makecell{0.848$_{\pm 0.045}$ \\ 0.569$_{\pm 0.019}$}} & 
        \multicolumn{1}{c}{\cellcolor{r1} \makecell{0.875$_{\pm 0.015}$ \\ 0.630$_{\pm 0.047}$}} \\
        \midrule
        \makecell{BatteryBERT \\ \citep{huang2022batterybert}} & 
        \multicolumn{1}{c}{\makecell{0.594$_{\pm 0.085}$ \\ 0.359$_{\pm 0.075}$}} &
        \multicolumn{1}{c}{\makecell{0.592$_{\pm 0.084}$ \\ 0.297$_{\pm 0.025}$}} & 
        \multicolumn{1}{c}{\makecell{0.423$_{\pm 0.097}$ \\ 0.167$_{\pm 0.074}$}} & 
        \multicolumn{1}{c}{\makecell{0.823$_{\pm 0.073}$ \\ 0.553$_{\pm 0.074}$}} & 
        \multicolumn{1}{c}{\makecell{0.801$_{\pm 0.081}$ \\ 0.466$_{\pm 0.111}$}} & 
        \multicolumn{1}{c}{\cellcolor{r1} \makecell{0.854$_{\pm 0.029}$ \\ 0.592$_{\pm 0.066}$}} & 
        \multicolumn{1}{c}{\makecell{0.786$_{\pm 0.113}$ \\ 0.472$_{\pm 0.150}$}} \\
        \midrule
        \makecell{SciBERT \\ \citep{beltagy2019scibert}} & 
        \multicolumn{1}{c}{\makecell{0.699$_{\pm 0.105}$ \\ 0.495$_{\pm 0.099}$}} &
        \multicolumn{1}{c}{\makecell{0.585$_{\pm 0.125}$ \\ 0.267$_{\pm 0.042}$}} & 
        \multicolumn{1}{c}{\makecell{0.643$_{\pm 0.088}$ \\ 0.311$_{\pm 0.098}$}} & 
        \multicolumn{1}{c}{\makecell{0.799$_{\pm 0.139}$ \\ 0.527$_{\pm 0.204}$}} & 
        \multicolumn{1}{c}{\makecell{0.783$_{\pm 0.085}$ \\ 0.474$_{\pm 0.099}$}} & 
        \multicolumn{1}{c}{\makecell{0.814$_{\pm 0.125}$ \\ 0.528$_{\pm 0.180}$}} & 
        \multicolumn{1}{c}{\cellcolor{r1} \makecell{0.819$_{\pm 0.067}$ \\ 0.545$_{\pm 0.119}$}} \\
        \midrule
        \makecell{ScholarBERT \\ \citep{hong2022scholarbert}} & 
        \multicolumn{1}{c}{\makecell{0.603$_{\pm 0.179}$ \\ 0.178$_{\pm 0.186}$}} &
        \multicolumn{1}{c}{\cellcolor{r1} \makecell{0.619$_{\pm 0.248}$ \\ 0.384$_{\pm 0.154}$}} & 
        \multicolumn{1}{c}{\makecell{0.243$_{\pm 0.351}$ \\ 0.078$_{\pm 0.139}$}} & 
        \multicolumn{1}{c}{\makecell{0.416$_{\pm 0.013}$ \\ 0.334$_{\pm 0.006}$}} & 
        \multicolumn{1}{c}{\makecell{0.543$_{\pm 0.060}$ \\ 0.252$_{\pm 0.062}$}} & 
        \multicolumn{1}{c}{\makecell{0.367$_{\pm 0.080}$ \\ 0.236$_{\pm 0.119}$}} & 
        \multicolumn{1}{c}{\makecell{0.428$_{\pm 0.148}$ \\ 0.274$_{\pm 0.110}$}} \\
        \midrule
        \makecell{BioBERT \\ \citep{shoyabiobert}} & 
        \multicolumn{1}{c}{\makecell{0.692$_{\pm 0.105}$ \\ 0.458$_{\pm 0.087}$}} &
        \multicolumn{1}{c}{\makecell{0.538$_{\pm 0.108}$ \\ 0.243$_{\pm 0.029}$}} & 
        \multicolumn{1}{c}{\makecell{0.306$_{\pm 0.032}$ \\ 0.079$_{\pm 0.017}$}} & 
        \multicolumn{1}{c}{\makecell{0.743$_{\pm 0.199}$ \\ 0.442$_{\pm 0.215}$}} & 
        \multicolumn{1}{c}{\makecell{0.674$_{\pm 0.093}$ \\ 0.323$_{\pm 0.092}$}} & 
        \multicolumn{1}{c}{\makecell{0.666$_{\pm 0.220}$ \\ 0.324$_{\pm 0.118}$}} & 
        \multicolumn{1}{c}{\cellcolor{r1} \makecell{0.797$_{\pm 0.092}$ \\ 0.465$_{\pm 0.134}$}} \\
        \midrule
        \makecell{BERT \\ \citep{devlin2018bert}} & 
        \multicolumn{1}{c}{\makecell{0.564$_{\pm 0.130}$ \\ 0.357$_{\pm 0.076}$}} &
        \multicolumn{1}{c}{\makecell{0.626$_{\pm 0.103}$ \\ 0.306$_{\pm 0.075}$}} & 
        \multicolumn{1}{c}{\makecell{0.368$_{\pm 0.112}$ \\ 0.100$_{\pm 0.018}$}} & 
        \multicolumn{1}{c}{\cellcolor{r1} \makecell{0.792$_{\pm 0.056}$ \\ 0.533$_{\pm 0.041}$}} & 
        \multicolumn{1}{c}{\makecell{0.696$_{\pm 0.046}$ \\ 0.382$_{\pm 0.039}$}} & 
        \multicolumn{1}{c}{\makecell{0.636$_{\pm 0.094}$ \\ 0.382$_{\pm 0.043}$}} & 
        \multicolumn{1}{c}{\makecell{0.782$_{\pm 0.056}$ \\ 0.494$_{\pm 0.061}$}} \\
        \bottomrule
    \end{tabular}
\end{adjustbox}
\end{spacing}
\vspace{-2mm}
\caption{Results of \textbf{relation classification} task among seven tasks on different schema settings for various BERT models pre-trained on different domain specific text data. For each model, the top line represents the micro-F1 score and the bottom line represents the macro-F1 score. We report the mean across 5 experiments with a confidence interval of two standard deviations. We highlight the \hlc[r1]{best} performing method.}
\label{table:schema-re}
\end{table*}

\begin{table*}[h]
\begin{spacing}{1.5}
\centering

\vspace{-2.5mm}
\begin{adjustbox}{max width=1\linewidth}
    \begin{tabular}{c|c|c|c|c|c|c|c}
        \toprule
        \bf{NLP Model} & 
        \multicolumn{1}{c}{\makecell{\bf{Single Task}}} & 
        \multicolumn{1}{c}{\makecell{\bf{Single Task Prompt}}} &
        \multicolumn{1}{c}{\makecell{\bf{MMOE}}} & 
        \multicolumn{1}{c}{\makecell{\bf{No Explanations}}} & 
        \multicolumn{1}{c}{\makecell{\bf{Potential} \\ \bf{Choices}}} & 
        \multicolumn{1}{c}{\makecell{\bf{Examples}}} & 
        \multicolumn{1}{c}{\makecell{\bf{Text2Schema}}} \\
        \midrule
        \makecell{MatSciBERT \\ \citep{gupta2022matscibert}} & 
        \multicolumn{1}{c}{\makecell{0.108$_{\pm 0.062}$ \\ 0.041$_{\pm 0.020}$}} & 
        \multicolumn{1}{c}{\makecell{0.148$_{\pm 0.182}$ \\ 0.050$_{\pm 0.071}$}} & 
        \multicolumn{1}{c}{\makecell{0.280$_{\pm 0.127}$ \\ 0.122$_{\pm 0.063}$}} &
        \multicolumn{1}{c}{\makecell{0.448$_{\pm 0.091}$ \\ 0.251$_{\pm 0.075}$}} & 
        \multicolumn{1}{c}{\cellcolor{r1} \makecell{0.498$_{\pm 0.045}$ \\ 0.310$_{\pm 0.036}$}} & 
        \multicolumn{1}{c}{\makecell{0.484$_{\pm 0.015}$ \\ 0.292$_{\pm 0.052}$}} & 
        \multicolumn{1}{c}{\makecell{0.436$_{\pm 0.066}$ \\ 0.251$_{\pm 0.075}$}} \\
        \midrule
        \makecell{MatBERT \\ \citep{walker2021impact}} &  
        \multicolumn{1}{c}{\makecell{0.152$_{\pm 0.093}$ \\ 0.029$_{\pm 0.021}$}} & 
        \multicolumn{1}{c}{\makecell{0.160$_{\pm 0.169}$ \\ 0.033$_{\pm 0.033}$}} & 
        \multicolumn{1}{c}{\makecell{0.341$_{\pm 0.006}$ \\ 0.174$_{\pm 0.027}$}} &
        \multicolumn{1}{c}{\makecell{0.453$_{\pm 0.108}$ \\ 0.274$_{\pm 0.087}$}} & 
        \multicolumn{1}{c}{\makecell{0.483$_{\pm 0.063}$ \\ 0.298$_{\pm 0.037}$}} & 
        \multicolumn{1}{c}{\cellcolor{r1} \makecell{0.515$_{\pm 0.040}$ \\ 0.288$_{\pm 0.064}$}} & 
        \multicolumn{1}{c}{\makecell{0.451$_{\pm 0.091}$ \\ 0.288$_{\pm 0.066}$}} \\
        \midrule
        \makecell{BatteryBERT \\ \citep{huang2022batterybert}} &  
        \multicolumn{1}{c}{\makecell{0.149$_{\pm 0.072}$ \\ 0.030$_{\pm 0.039}$}} & 
        \multicolumn{1}{c}{\makecell{0.162$_{\pm 0.166}$ \\ 0.036$_{\pm 0.029}$}} & 
        \multicolumn{1}{c}{\makecell{0.232$_{\pm 0.196}$ \\ 0.104$_{\pm 0.088}$}} &
        \multicolumn{1}{c}{\makecell{0.397$_{\pm 0.105}$ \\ 0.233$_{\pm 0.086}$}} & 
        \multicolumn{1}{c}{\makecell{0.438$_{\pm 0.063}$ \\ 0.298$_{\pm 0.037}$}} & 
        \multicolumn{1}{c}{\makecell{0.443$_{\pm 0.023}$ \\ 0.250$_{\pm 0.068}$}} & 
        \multicolumn{1}{c}{\cellcolor{r1} \makecell{0.457$_{\pm 0.024}$ \\ 0.277$_{\pm 0.034}$}} \\
        \midrule
        \makecell{SciBERT \\ \citep{beltagy2019scibert}} & 
        \multicolumn{1}{c}{\makecell{0.152$_{\pm 0.123}$ \\ 0.041$_{\pm 0.068}$}} & 
        \multicolumn{1}{c}{\makecell{0.160$_{\pm 0.189}$ \\ 0.033$_{\pm 0.032}$}} & 
        \multicolumn{1}{c}{\makecell{0.312$_{\pm 0.015}$ \\ 0.159$_{\pm 0.024}$}} &
        \multicolumn{1}{c}{\makecell{0.449$_{\pm 0.079}$ \\ 0.259$_{\pm 0.072}$}} & 
        \multicolumn{1}{c}{\makecell{0.442$_{\pm 0.135}$ \\ 0.264$_{\pm 0.103}$}} & 
        \multicolumn{1}{c}{\cellcolor{r1} \makecell{0.484$_{\pm 0.042}$ \\ 0.287$_{\pm 0.075}$}} & 
        \multicolumn{1}{c}{\makecell{0.451$_{\pm 0.077}$ \\ 0.276$_{\pm 0.080}$}} \\
        \midrule
        \makecell{ScholarBERT \\ \citep{hong2022scholarbert}} & 
        \multicolumn{1}{c}{\makecell{0.349$_{\pm 0.102}$ \\ 0.250$_{\pm 0.101}$}} & 
        \multicolumn{1}{c}{\makecell{0.444$_{\pm 0.091}$ \\ 0.253$_{\pm 0.103}$}} & 
        \multicolumn{1}{c}{\makecell{0.262$_{\pm 0.062}$ \\ 0.102$_{\pm 0.108}$}} &
        \multicolumn{1}{c}{\makecell{0.454$_{\pm 0.094}$ \\ 0.312$_{\pm 0.131}$}} & 
        \multicolumn{1}{c}{\makecell{0.454$_{\pm 0.095}$ \\ 0.264$_{\pm 0.102}$}} & 
        \multicolumn{1}{c}{\makecell{0.431$_{\pm 0.081}$ \\ 0.296$_{\pm 0.144}$}} & 
        \multicolumn{1}{c}{\cellcolor{r1} \makecell{0.489$_{\pm 0.083}$ \\ 0.356$_{\pm 0.109}$}} \\
        \midrule
        \makecell{BioBERT \\ \citep{shoyabiobert}} & 
        \multicolumn{1}{c}{\makecell{0.119$_{\pm 0.080}$ \\ 0.030$_{\pm 0.011}$}} & 
        \multicolumn{1}{c}{\makecell{0.160$_{\pm 0.170}$ \\ 0.034$_{\pm 0.032}$}} & 
        \multicolumn{1}{c}{\makecell{0.054$_{\pm 0.000}$ \\ 0.013$_{\pm 0.000}$}} &
        \multicolumn{1}{c}{\makecell{0.489$_{\pm 0.058}$ \\ 0.305$_{\pm 0.090}$}} & 
        \multicolumn{1}{c}{\cellcolor{r1} \makecell{0.491$_{\pm 0.027}$ \\ 0.295$_{\pm 0.059}$}} & 
        \multicolumn{1}{c}{\makecell{0.473$_{\pm 0.034}$ \\ 0.268$_{\pm 0.061}$}} & 
        \multicolumn{1}{c}{\makecell{0.488$_{\pm 0.036}$ \\ 0.274$_{\pm 0.049}$}} \\
        \midrule
        \makecell{BERT \\ \citep{devlin2018bert}} & 
        \multicolumn{1}{c}{\makecell{0.198$_{\pm 0.041}$ \\ 0.042$_{\pm 0.055}$}} & 
        \multicolumn{1}{c}{\makecell{0.160$_{\pm 0.170}$ \\ 0.033$_{\pm 0.033}$}} & 
        \multicolumn{1}{c}{\makecell{0.232$_{\pm 0.002}$ \\ 0.049$_{\pm 0.008}$}} & 
        \multicolumn{1}{c}{\makecell{0.400$_{\pm 0.017}$ \\ 0.194$_{\pm 0.025}$}} & 
        \multicolumn{1}{c}{\makecell{0.414$_{\pm 0.064}$ \\ 0.214$_{\pm 0.092}$}} & 
        \multicolumn{1}{c}{\cellcolor{r1} \makecell{0.451$_{\pm 0.074}$ \\ 0.265$_{\pm 0.104}$}} & 
        \multicolumn{1}{c}{\makecell{0.418$_{\pm 0.053}$ \\ 0.225$_{\pm 0.091}$}} \\
        \bottomrule
    \end{tabular}
\end{adjustbox}
\end{spacing}
\vspace{-2mm}
\caption{Results of \textbf{event argument extraction} task among seven tasks on different schema settings for various BERT models pre-trained on different domain specific text data. For each model, the top line represents the micro-F1 score and the bottom line represents the macro-F1 score. We report the mean across 5 experiments with a confidence interval of two standard deviations. We highlight the \hlc[r1]{best} performing method.}
\label{table:schema-ee}
\end{table*}

\begin{table*}[h]
\begin{spacing}{1.5}
\centering

\vspace{-2.5mm}
\begin{adjustbox}{max width=1\linewidth}
    \begin{tabular}{c|c|c|c|c|c|c|c}
        \toprule
        \bf{NLP Model} & 
        \multicolumn{1}{c}{\makecell{\bf{Single Task}}} & 
        \multicolumn{1}{c}{\makecell{\bf{Single Task Prompt}}} &
        \multicolumn{1}{c}{\makecell{\bf{MMOE}}} & 
        \multicolumn{1}{c}{\makecell{\bf{No Explanations}}} & 
        \multicolumn{1}{c}{\makecell{\bf{Potential} \\ \bf{Choices}}} & 
        \multicolumn{1}{c}{\makecell{\bf{Examples}}} & 
        \multicolumn{1}{c}{\makecell{\bf{Text2Schema}}} \\
        \midrule
        \makecell{MatSciBERT \\ \citep{gupta2022matscibert}} & 
        \multicolumn{1}{c}{\makecell{0.685$_{\pm 0.074}$ \\ 0.588$_{\pm 0.152}$}} & 
        \multicolumn{1}{c}{\makecell{0.673$_{\pm 0.003}$ \\ 0.402$_{\pm 0.001}$}} & 
        \multicolumn{1}{c}{\makecell{0.607$_{\pm 0.277}$ \\ 0.386$_{\pm 0.150}$}} & 
        \multicolumn{1}{c}{\makecell{0.706$_{\pm 0.013}$ \\ 0.633$_{\pm 0.115}$}} & 
        \multicolumn{1}{c}{\makecell{0.694$_{\pm 0.041}$ \\ 0.524$_{\pm 0.175}$}} & 
        \multicolumn{1}{c}{\makecell{0.686$_{\pm 0.158}$ \\ 0.583$_{\pm 0.226}$}} & 
        \multicolumn{1}{c}{\cellcolor{r1} \makecell{0.719$_{\pm 0.116}$ \\ 0.623$_{\pm 0.183}$}} \\
        \midrule
        \makecell{MatBERT \\ \citep{walker2021impact}} &  
        \multicolumn{1}{c}{\makecell{0.753$_{\pm 0.031}$ \\ 0.730$_{\pm 0.016}$}} & 
        \multicolumn{1}{c}{\makecell{0.671$_{\pm 0.002}$ \\ 0.402$_{\pm 0.001}$}} & 
        \multicolumn{1}{c}{\makecell{0.673$_{\pm 0.001}$ \\ 0.404$_{\pm 0.004}$}} & 
        \multicolumn{1}{c}{\makecell{0.727$_{\pm 0.089}$ \\ 0.601$_{\pm 0.212}$}} & 
        \multicolumn{1}{c}{\cellcolor{r1} \makecell{0.776$_{\pm 0.059}$ \\ 0.722$_{\pm 0.076}$}} & 
        \multicolumn{1}{c}{\makecell{0.649$_{\pm 0.039}$ \\ 0.509$_{\pm 0.155}$}} & 
        \multicolumn{1}{c}{\makecell{0.756$_{\pm 0.073}$ \\ 0.691$_{\pm 0.188}$}} \\
        \midrule
        \makecell{BatteryBERT \\ \citep{huang2022batterybert}} &  
        \multicolumn{1}{c}{\makecell{0.663$_{\pm 0.088}$ \\ 0.585$_{\pm 0.156}$}} & 
        \multicolumn{1}{c}{\cellcolor{r1} \makecell{0.672$_{\pm 0.001}$ \\ 0.402$_{\pm 0.000}$}} & 
        \multicolumn{1}{c}{\cellcolor{r1} \makecell{0.672$_{\pm 0.002}$ \\ 0.402$_{\pm 0.001}$}} & 
        \multicolumn{1}{c}{\makecell{0.621$_{\pm 0.160}$ \\ 0.564$_{\pm 0.180}$}} & 
        \multicolumn{1}{c}{\makecell{0.626$_{\pm 0.113}$ \\ 0.574$_{\pm 0.092}$}} & 
        \multicolumn{1}{c}{\cellcolor{r1} \makecell{0.672$_{\pm 0.031}$ \\ 0.540$_{\pm 0.129}$}} & 
        \multicolumn{1}{c}{\makecell{0.633$_{\pm 0.075}$ \\ 0.610$_{\pm 0.046}$}} \\
        \midrule
        \makecell{SciBERT \\ \citep{beltagy2019scibert}} & 
        \multicolumn{1}{c}{\makecell{0.690$_{\pm 0.074}$ \\ 0.605$_{\pm 0.150}$}} & 
        \multicolumn{1}{c}{\makecell{0.673$_{\pm 0.002}$ \\ 0.402$_{\pm 0.001}$}} & 
        \multicolumn{1}{c}{\makecell{0.568$_{\pm 0.289}$ \\ 0.370$_{\pm 0.089}$}} & 
        \multicolumn{1}{c}{\makecell{0.703$_{\pm 0.041}$ \\ 0.598$_{\pm 0.204}$}} & 
        \multicolumn{1}{c}{\cellcolor{r1} \makecell{0.711$_{\pm 0.076}$ \\ 0.598$_{\pm 0.203}$}} & 
        \multicolumn{1}{c}{\makecell{0.662$_{\pm 0.169}$ \\ 0.562$_{\pm 0.202}$}} & 
        \multicolumn{1}{c}{\makecell{0.696$_{\pm 0.094}$ \\ 0.546$_{\pm 0.243}$}} \\
        \midrule
        \makecell{ScholarBERT \\ \citep{hong2022scholarbert}} & 
        \multicolumn{1}{c}{\makecell{0.620$_{\pm 0.161}$ \\ 0.386$_{\pm 0.150}$}} & 
        \multicolumn{1}{c}{\makecell{0.603$_{\pm 0.271}$ \\ 0.371$_{\pm 0.122}$}} & 
        \multicolumn{1}{c}{\makecell{0.658$_{\pm 0.029}$ \\ 0.407$_{\pm 0.010}$}} & 
        \multicolumn{1}{c}{\cellcolor{r1} \makecell{0.672$_{\pm 0.003}$ \\ 0.482$_{\pm 0.001}$}} & 
        \multicolumn{1}{c}{\makecell{0.662$_{\pm 0.144}$ \\ 0.534$_{\pm 0.260}$}} & 
        \multicolumn{1}{c}{\makecell{0.668$_{\pm 0.016}$ \\ 0.405$_{\pm 0.007}$}} & 
        \multicolumn{1}{c}{\makecell{0.663$_{\pm 0.032}$ \\ 0.433$_{\pm 0.122}$}} \\
        \midrule
        \makecell{BioBERT \\ \citep{shoyabiobert}} & 
        \multicolumn{1}{c}{\makecell{0.629$_{\pm 0.041}$ \\ 0.507$_{\pm 0.033}$}} & 
        \multicolumn{1}{c}{\makecell{0.672$_{\pm 0.002}$ \\ 0.402$_{\pm 0.001}$}} & 
        \multicolumn{1}{c}{\makecell{0.671$_{\pm 0.001}$ \\ 0.401$_{\pm 0.001}$}} & 
        \multicolumn{1}{c}{\makecell{0.658$_{\pm 0.211}$ \\ 0.588$_{\pm 0.258}$}} & 
        \multicolumn{1}{c}{\cellcolor{r1} \makecell{0.709$_{\pm 0.033}$ \\ 0.651$_{\pm 0.081}$}} & 
        \multicolumn{1}{c}{\makecell{0.680$_{\pm 0.193}$ \\ 0.622$_{\pm 0.226}$}} & 
        \multicolumn{1}{c}{\makecell{0.675$_{\pm 0.144}$ \\ 0.578$_{\pm 0.102}$}} \\
        \midrule
        \makecell{BERT \\ \citep{devlin2018bert}} & 
        \multicolumn{1}{c}{\makecell{0.709$_{\pm 0.090}$ \\ 0.585$_{\pm 0.093}$}} & 
        \multicolumn{1}{c}{\makecell{0.672$_{\pm 0.001}$ \\ 0.468$_{\pm 0.283}$}} & 
        \multicolumn{1}{c}{\makecell{0.672$_{\pm 0.003}$ \\ 0.402$_{\pm 0.001}$}} & 
        \multicolumn{1}{c}{\makecell{0.685$_{\pm 0.050}$ \\ 0.562$_{\pm 0.221}$}} & 
        \multicolumn{1}{c}{\cellcolor{r1} \makecell{0.727$_{\pm 0.102}$ \\ 0.602$_{\pm 0.238}$}} & 
        \multicolumn{1}{c}{\makecell{0.629$_{\pm 0.291}$ \\ 0.468$_{\pm 0.283}$}} & 
        \multicolumn{1}{c}{\makecell{0.665$_{\pm 0.057}$ \\ 0.532$_{\pm 0.194}$}} \\
        \bottomrule
    \end{tabular}
\end{adjustbox}
\end{spacing}
\vspace{-2mm}
\caption{Results of \textbf{paragraph classification} task among seven tasks on different schema settings for various BERT models pre-trained on different domain specific text data. For each model, the top line represents the micro-F1 score and the bottom line represents the macro-F1 score. We report the mean across 5 experiments with a confidence interval of two standard deviations. We highlight the \hlc[r1]{best} performing method.}
\label{table:schema-pc}
\end{table*}

\begin{table*}[h]
\begin{spacing}{1.5}
\centering

\vspace{-2.5mm}
\begin{adjustbox}{max width=1\linewidth}
    \begin{tabular}{c|c|c|c|c|c|c|c}
        \toprule
        \bf{NLP Model} & 
        \multicolumn{1}{c}{\makecell{\bf{Single Task}}} & 
        \multicolumn{1}{c}{\makecell{\bf{Single Task Prompt}}} &
        \multicolumn{1}{c}{\makecell{\bf{MMOE}}} & 
        \multicolumn{1}{c}{\makecell{\bf{No Explanations}}} & 
        \multicolumn{1}{c}{\makecell{\bf{Potential} \\ \bf{Choices}}} & 
        \multicolumn{1}{c}{\makecell{\bf{Examples}}} & 
        \multicolumn{1}{c}{\makecell{\bf{Text2Schema}}} \\
        \midrule
        \makecell{MatSciBERT \\ \citep{gupta2022matscibert}} & 
        \multicolumn{1}{c}{\makecell{0.383$_{\pm 0.024}$ \\ 0.082$_{\pm 0.009}$}} & 
        \multicolumn{1}{c}{\makecell{0.334$_{\pm 0.004}$ \\ 0.063$_{\pm 0.001}$}} & 
        \multicolumn{1}{c}{\makecell{0.424$_{\pm 0.249}$ \\ 0.169$_{\pm 0.096}$}} &
        \multicolumn{1}{c}{\makecell{0.676$_{\pm 0.071}$ \\ 0.505$_{\pm 0.094}$}} & 
        \multicolumn{1}{c}{\makecell{0.631$_{\pm 0.081}$ \\ 0.445$_{\pm 0.153}$}} & 
        \multicolumn{1}{c}{\cellcolor{r1} \makecell{0.741$_{\pm 0.157}$ \\ 0.549$_{\pm 0.179}$}} & 
        \multicolumn{1}{c}{\makecell{0.692$_{\pm 0.179}$ \\ 0.484$_{\pm 0.254}$}} \\
        \midrule
        \makecell{MatBERT \\ \citep{walker2021impact}} &  
        \multicolumn{1}{c}{\makecell{0.346$_{\pm 0.006}$ \\ 0.067$_{\pm 0.004}$}} & 
        \multicolumn{1}{c}{\makecell{0.334$_{\pm 0.001}$ \\ 0.063$_{\pm 0.000}$}} & 
        \multicolumn{1}{c}{\makecell{0.549$_{\pm 0.087}$ \\ 0.300$_{\pm 0.045}$}} &
        \multicolumn{1}{c}{\cellcolor{r1} \makecell{0.792$_{\pm 0.073}$ \\ 0.653$_{\pm 0.184}$}} & 
        \multicolumn{1}{c}{\makecell{0.669$_{\pm 0.061}$ \\ 0.497$_{\pm 0.086}$}} & 
        \multicolumn{1}{c}{\makecell{0.744$_{\pm 0.010}$ \\ 0.557$_{\pm 0.082}$}} & 
        \multicolumn{1}{c}{\makecell{0.717$_{\pm 0.040}$ \\ 0.549$_{\pm 0.091}$}} \\
        \midrule
        \makecell{BatteryBERT \\ \citep{huang2022batterybert}} &  
        \multicolumn{1}{c}{\makecell{0.280$_{\pm 0.004}$ \\ 0.118$_{\pm 0.041}$}} & 
        \multicolumn{1}{c}{\makecell{0.334$_{\pm 0.001}$ \\ 0.063$_{\pm 0.000}$}} & 
        \multicolumn{1}{c}{\makecell{0.311$_{\pm 0.062}$ \\ 0.073$_{\pm 0.028}$}} &
        \multicolumn{1}{c}{\cellcolor{r1} \makecell{0.670$_{\pm 0.046}$ \\ 0.496$_{\pm 0.117}$}} & 
        \multicolumn{1}{c}{\makecell{0.558$_{\pm 0.179}$ \\ 0.358$_{\pm 0.149}$}} & 
        \multicolumn{1}{c}{\makecell{0.492$_{\pm 0.181}$ \\ 0.282$_{\pm 0.184}$}} & 
        \multicolumn{1}{c}{\makecell{0.614$_{\pm 0.128}$ \\ 0.419$_{\pm 0.149}$}} \\
        \midrule
        \makecell{SciBERT \\ \citep{beltagy2019scibert}} & 
        \multicolumn{1}{c}{\makecell{0.281$_{\pm 0.009}$ \\ 0.052$_{\pm 0.027}$}} & 
        \multicolumn{1}{c}{\makecell{0.334$_{\pm 0.001}$ \\ 0.063$_{\pm 0.001}$}} & 
        \multicolumn{1}{c}{\makecell{0.455$_{\pm 0.081}$ \\ 0.207$_{\pm 0.095}$}} &
        \multicolumn{1}{c}{\makecell{0.727$_{\pm 0.114}$ \\ 0.564$_{\pm 0.137}$}} & 
        \multicolumn{1}{c}{\makecell{0.623$_{\pm 0.069}$ \\ 0.456$_{\pm 0.135}$}} & 
        \multicolumn{1}{c}{\cellcolor{r1} \makecell{0.740$_{\pm 0.133}$ \\ 0.533$_{\pm 0.160}$}} & 
        \multicolumn{1}{c}{\makecell{0.701$_{\pm 0.138}$ \\ 0.516$_{\pm 0.217}$}} \\
        \midrule
        \makecell{ScholarBERT \\ \citep{hong2022scholarbert}} & 
        \multicolumn{1}{c}{\makecell{0.437$_{\pm 0.104}$ \\ 0.193$_{\pm 0.076}$}} & 
        \multicolumn{1}{c}{\makecell{0.489$_{\pm 0.105}$ \\ 0.266$_{\pm 0.105}$}} & 
        \multicolumn{1}{c}{\makecell{0.330$_{\pm 0.007}$ \\ 0.070$_{\pm 0.015}$}} &
        \multicolumn{1}{c}{\makecell{0.389$_{\pm 0.001}$ \\ 0.190$_{\pm 0.000}$}} & 
        \multicolumn{1}{c}{\cellcolor{r1} \makecell{0.492$_{\pm 0.165}$ \\ 0.308$_{\pm 0.156}$}} & 
        \multicolumn{1}{c}{\makecell{0.389$_{\pm 0.001}$ \\ 0.191$_{\pm 0.001}$}} & 
        \multicolumn{1}{c}{\makecell{0.322$_{\pm 0.260}$ \\ 0.178$_{\pm 0.051}$}} \\
        \midrule
        \makecell{BioBERT \\ \citep{shoyabiobert}} & 
        \multicolumn{1}{c}{\makecell{0.300$_{\pm 0.015}$ \\ 0.073$_{\pm 0.002}$}} & 
        \multicolumn{1}{c}{\makecell{0.324$_{\pm 0.001}$ \\ 0.062$_{\pm 0.000}$}} & 
        \multicolumn{1}{c}{\makecell{0.334$_{\pm 0.062}$ \\ 0.073$_{\pm 0.027}$}} &
        \multicolumn{1}{c}{\cellcolor{r1} \makecell{0.662$_{\pm 0.060}$ \\ 0.426$_{\pm 0.078}$}} & 
        \multicolumn{1}{c}{\makecell{0.561$_{\pm 0.128}$ \\ 0.346$_{\pm 0.133}$}} & 
        \multicolumn{1}{c}{\makecell{0.545$_{\pm 0.157}$ \\ 0.347$_{\pm 0.128}$}} & 
        \multicolumn{1}{c}{\makecell{0.647$_{\pm 0.140}$ \\ 0.446$_{\pm 0.231}$}} \\
        \midrule
        \makecell{BERT \\ \citep{devlin2018bert}} & 
        \multicolumn{1}{c}{\makecell{0.348$_{\pm 0.047}$ \\ 0.091$_{\pm 0.020}$}} & 
        \multicolumn{1}{c}{\makecell{0.334$_{\pm 0.001}$ \\ 0.063$_{\pm 0.000}$}} & 
        \multicolumn{1}{c}{\makecell{0.313$_{\pm 0.083}$ \\ 0.073$_{\pm 0.037}$}} & 
        \multicolumn{1}{c}{\cellcolor{r1} \makecell{0.668$_{\pm 0.061}$ \\ 0.495$_{\pm 0.058}$}} & 
        \multicolumn{1}{c}{\makecell{0.593$_{\pm 0.059}$ \\ 0.424$_{\pm 0.086}$}} & 
        \multicolumn{1}{c}{\makecell{0.594$_{\pm 0.081}$ \\ 0.371$_{\pm 0.103}$}} & 
        \multicolumn{1}{c}{\makecell{0.656$_{\pm 0.099}$ \\ 0.515$_{\pm 0.067}$}} \\
        \bottomrule
    \end{tabular}
\end{adjustbox}
\end{spacing}
\vspace{-2mm}
\caption{Results of \textbf{synthesis action retrieval} task among seven tasks on different schema settings for various BERT models pre-trained on different domain specific text data. For each model, the top line represents the micro-F1 score and the bottom line represents the macro-F1 score. We report the mean across 5 experiments with a confidence interval of two standard deviations. We highlight the \hlc[r1]{best} performing method.}
\label{table:schema-sar}
\end{table*}

\begin{table*}[h]
\begin{spacing}{1.5}
\centering

\vspace{-2.5mm}
\begin{adjustbox}{max width=1\linewidth}
    \begin{tabular}{c|c|c|c|c|c|c|c}
        \toprule
        \bf{NLP Model} & 
        \multicolumn{1}{c}{\makecell{\bf{Single Task}}} & 
        \multicolumn{1}{c}{\makecell{\bf{Single Task Prompt}}} & 
        \multicolumn{1}{c}{\makecell{\bf{MMOE}}} & 
        \multicolumn{1}{c}{\makecell{\bf{No Explanations}}} & 
        \multicolumn{1}{c}{\makecell{\bf{Potential} \\ \bf{Choices}}} & 
        \multicolumn{1}{c}{\makecell{\bf{Examples}}} & 
        \multicolumn{1}{c}{\makecell{\bf{Text2Schema}}} \\
        \midrule
        \makecell{MatSciBERT \\ \citep{gupta2022matscibert}} & 
        \multicolumn{1}{c}{\makecell{0.888$_{\pm 0.093}$ \\ 0.602$_{\pm 0.151}$}} & 
        \multicolumn{1}{c}{\makecell{0.908$_{\pm 0.001}$ \\ 0.476$_{\pm 0.001}$}} & 
        \multicolumn{1}{c}{\makecell{0.907$_{\pm 0.001}$ \\ 0.493$_{\pm 0.069}$}} & 
        \multicolumn{1}{c}{\makecell{0.908$_{\pm 0.010}$ \\ 0.601$_{\pm 0.159}$}} & 
        \multicolumn{1}{c}{\makecell{0.903$_{\pm 0.019}$ \\ 0.573$_{\pm 0.135}$}} & 
        \multicolumn{1}{c}{\makecell{0.905$_{\pm 0.020}$ \\ 0.616$_{\pm 0.150}$}} & 
        \multicolumn{1}{c}{\cellcolor{r1} \makecell{0.914$_{\pm 0.008}$ \\ 0.660$_{\pm 0.079}$}} \\
        \midrule
        \makecell{MatBERT \\ \citep{walker2021impact}} &  
        \multicolumn{1}{c}{\makecell{0.908$_{\pm 0.011}$ \\ 0.441$_{\pm 0.038}$}} & 
        \multicolumn{1}{c}{\makecell{0.908$_{\pm 0.001}$ \\ 0.476$_{\pm 0.001}$}} & 
        \multicolumn{1}{c}{\makecell{0.907$_{\pm 0.000}$ \\ 0.476$_{\pm 0.000}$}} & 
        \multicolumn{1}{c}{\makecell{0.906$_{\pm 0.016}$ \\ 0.645$_{\pm 0.025}$}} & 
        \multicolumn{1}{c}{\cellcolor{r1} \makecell{0.910$_{\pm 0.012}$ \\ 0.561$_{\pm 0.135}$}} & 
        \multicolumn{1}{c}{\makecell{0.903$_{\pm 0.018}$ \\ 0.600$_{\pm 0.089}$}} & 
        \multicolumn{1}{c}{\makecell{0.909$_{\pm 0.009}$ \\ 0.614$_{\pm 0.134}$}} \\
        \midrule
        \makecell{BatteryBERT \\ \citep{huang2022batterybert}} &  
        \multicolumn{1}{c}{\makecell{0.908$_{\pm 0.012}$ \\ 0.452$_{\pm 0.045}$}} & 
        \multicolumn{1}{c}{\makecell{0.907$_{\pm 0.000}$ \\ 0.475$_{\pm 0.001}$}} & 
        \multicolumn{1}{c}{\makecell{0.908$_{\pm 0.000}$ \\ 0.476$_{\pm 0.000}$}} & 
        \multicolumn{1}{c}{\makecell{0.895$_{\pm 0.050}$ \\ 0.679$_{\pm 0.080}$}} & 
        \multicolumn{1}{c}{\makecell{0.890$_{\pm 0.036}$ \\ 0.685$_{\pm 0.074}$}} & 
        \multicolumn{1}{c}{\makecell{0.907$_{\pm 0.002}$ \\ 0.519$_{\pm 0.144}$}} & 
        \multicolumn{1}{c}{\cellcolor{r1} \makecell{0.912$_{\pm 0.015}$ \\ 0.684$_{\pm 0.095}$}} \\
        \midrule
        \makecell{SciBERT \\ \citep{beltagy2019scibert}} & 
        \multicolumn{1}{c}{\makecell{0.896$_{\pm 0.080}$ \\ 0.421$_{\pm 0.159}$}} & 
        \multicolumn{1}{c}{\makecell{0.907$_{\pm 0.000}$ \\ 0.469$_{\pm 0.004}$}} & 
        \multicolumn{1}{c}{\makecell{0.825$_{\pm 0.218}$ \\ 0.535$_{\pm 0.079}$}} & 
        \multicolumn{1}{c}{\makecell{0.908$_{\pm 0.009}$ \\ 0.586$_{\pm 0.166}$}} & 
        \multicolumn{1}{c}{\makecell{0.902$_{\pm 0.017}$ \\ 0.596$_{\pm 0.161}$}} & 
        \multicolumn{1}{c}{\makecell{0.902$_{\pm 0.020}$ \\ 0.623$_{\pm 0.130}$}} & 
        \multicolumn{1}{c}{\cellcolor{r1} \makecell{0.911$_{\pm 0.017}$ \\ 0.617$_{\pm 0.143}$}} \\
        \midrule
        \makecell{ScholarBERT \\ \citep{hong2022scholarbert}} & 
        \multicolumn{1}{c}{\makecell{0.805$_{\pm 0.020}$ \\ 0.458$_{\pm 0.099}$}} & 
        \multicolumn{1}{c}{\makecell{0.839$_{\pm 0.268}$ \\ 0.477$_{\pm 0.004}$}} & 
        \multicolumn{1}{c}{\cellcolor{r1} \makecell{0.908$_{\pm 0.001}$ \\ 0.485$_{\pm 0.000}$}} & 
        \multicolumn{1}{c}{\cellcolor{r1} \makecell{0.908$_{\pm 0.000}$ \\ 0.476$_{\pm 0.000}$}} & 
        \multicolumn{1}{c}{\makecell{0.900$_{\pm 0.019}$ \\ 0.509$_{\pm 0.093}$}} & 
        \multicolumn{1}{c}{\makecell{0.907$_{\pm 0.001}$ \\ 0.476$_{\pm 0.001}$}} & 
        \multicolumn{1}{c}{\makecell{0.906$_{\pm 0.007}$ \\ 0.478$_{\pm 0.008}$}} \\
        \midrule
        \makecell{BioBERT \\ \citep{shoyabiobert}} & 
        \multicolumn{1}{c}{\makecell{0.908$_{\pm 0.001}$ \\ 0.476$_{\pm 0.001}$}} & 
        \multicolumn{1}{c}{\makecell{0.907$_{\pm 0.001}$ \\ 0.478$_{\pm 0.001}$}} & 
        \multicolumn{1}{c}{\makecell{0.907$_{\pm 0.001}$ \\ 0.503$_{\pm 0.005}$}} & 
        \multicolumn{1}{c}{\makecell{0.910$_{\pm 0.012}$ \\ 0.614$_{\pm 0.175}$}} & 
        \multicolumn{1}{c}{\makecell{0.899$_{\pm 0.047}$ \\ 0.610$_{\pm 0.078}$}} & 
        \multicolumn{1}{c}{\makecell{0.908$_{\pm 0.015}$ \\ 0.638$_{\pm 0.089}$}} & 
        \multicolumn{1}{c}{\cellcolor{r1} \makecell{0.915$_{\pm 0.021}$ \\ 0.686$_{\pm 0.098}$}} \\
        \midrule
        \makecell{BERT \\ \citep{devlin2018bert}} & 
        \multicolumn{1}{c}{\cellcolor{r1} \makecell{0.911$_{\pm 0.010}$ \\ 0.475$_{\pm 0.036}$}} & 
        \multicolumn{1}{c}{\makecell{0.907$_{\pm 0.000}$ \\ 0.476$_{\pm 0.000}$}} & 
        \multicolumn{1}{c}{\makecell{0.907$_{\pm 0.001}$ \\ 0.476$_{\pm 0.000}$}} & 
        \multicolumn{1}{c}{\makecell{0.906$_{\pm 0.007}$ \\ 0.549$_{\pm 0.086}$}} & 
        \multicolumn{1}{c}{\makecell{0.905$_{\pm 0.010}$ \\ 0.581$_{\pm 0.153}$}} & 
        \multicolumn{1}{c}{\makecell{0.892$_{\pm 0.035}$ \\ 0.563$_{\pm 0.136}$}} & 
        \multicolumn{1}{c}{\makecell{0.910$_{\pm 0.016}$ \\ 0.633$_{\pm 0.133}$}} \\
        \bottomrule
    \end{tabular}
\end{adjustbox}
\end{spacing}
\vspace{-2mm}
\caption{Results of \textbf{sentence classification} task among seven tasks on different schema settings for various BERT models pre-trained on different domain specific text data. For each model, the top line represents the micro-F1 score and the bottom line represents the macro-F1 score. We report the mean across 5 experiments with a confidence interval of two standard deviations. We highlight the \hlc[r1]{best} performing method.}
\label{table:schema-sc}
\end{table*}

\begin{table*}[h]
\begin{spacing}{1.5}
\centering

\vspace{-2.5mm}
\begin{adjustbox}{max width=1\linewidth}
    \begin{tabular}{c|c|c|c|c|c|c|c}
        \toprule
        \bf{NLP Model} & 
        \multicolumn{1}{c}{\makecell{\bf{Single Task}}} & 
        \multicolumn{1}{c}{\makecell{\bf{Single Task Prompt}}} &
        \multicolumn{1}{c}{\makecell{\bf{MMOE}}} & 
        \multicolumn{1}{c}{\makecell{\bf{No Explanations}}} & 
        \multicolumn{1}{c}{\makecell{\bf{Potential} \\ \bf{Choices}}} & 
        \multicolumn{1}{c}{\makecell{\bf{Examples}}} & 
        \multicolumn{1}{c}{\makecell{\bf{Text2Schema}}} \\
        \midrule
        \makecell{MatSciBERT \\ \citep{gupta2022matscibert}} & 
        \multicolumn{1}{c}{\makecell{0.083$_{\pm 0.047}$ \\ 0.087$_{\pm 0.045}$}} & 
        \multicolumn{1}{c}{\makecell{0.086$_{\pm 0.072}$ \\ 0.010$_{\pm 0.011}$}} & 
        \multicolumn{1}{c}{\makecell{0.043$_{\pm 0.023}$ \\ 0.016$_{\pm 0.005}$}} &
        \multicolumn{1}{c}{\makecell{0.419$_{\pm 0.074}$ \\ 0.182$_{\pm 0.043}$}} & 
        \multicolumn{1}{c}{\makecell{0.433$_{\pm 0.121}$ \\ 0.169$_{\pm 0.069}$}} & 
        \multicolumn{1}{c}{\makecell{0.428$_{\pm 0.187}$ \\ 0.169$_{\pm 0.075}$}} & 
        \multicolumn{1}{c}{\cellcolor{r1} \makecell{0.436$_{\pm 0.142}$ \\ 0.194$_{\pm 0.062}$}} \\
        \midrule
        \makecell{MatBERT \\ \citep{walker2021impact}} &  
        \multicolumn{1}{c}{\makecell{0.179$_{\pm 0.074}$ \\ 0.087$_{\pm 0.030}$}} & 
        \multicolumn{1}{c}{\makecell{0.151$_{\pm 0.121}$ \\ 0.024$_{\pm 0.022}$}} & 
        \multicolumn{1}{c}{\makecell{0.148$_{\pm 0.148}$ \\ 0.057$_{\pm 0.067}$}} &
        \multicolumn{1}{c}{\makecell{0.547$_{\pm 0.050}$ \\ 0.276$_{\pm 0.047}$}} & 
        \multicolumn{1}{c}{\makecell{0.493$_{\pm 0.078}$ \\ 0.230$_{\pm 0.067}$}} & 
        \multicolumn{1}{c}{\makecell{0.502$_{\pm 0.034}$ \\ 0.221$_{\pm 0.011}$}} & 
        \multicolumn{1}{c}{\cellcolor{r1} \makecell{0.548$_{\pm 0.058}$ \\ 0.273$_{\pm 0.051}$}} \\
        \midrule
        \makecell{BatteryBERT \\ \citep{huang2022batterybert}} &  
        \multicolumn{1}{c}{\makecell{0.093$_{\pm 0.074}$ \\ 0.009$_{\pm 0.012}$}} & 
        \multicolumn{1}{c}{\makecell{0.073$_{\pm 0.033}$ \\ 0.008$_{\pm 0.011}$}} & 
        \multicolumn{1}{c}{\makecell{0.032$_{\pm 0.031}$ \\ 0.008$_{\pm 0.009}$}} &
        \multicolumn{1}{c}{\cellcolor{r1} \makecell{0.540$_{\pm 0.092}$ \\ 0.270$_{\pm 0.108}$}} & 
        \multicolumn{1}{c}{\makecell{0.433$_{\pm 0.155}$ \\ 0.211$_{\pm 0.056}$}} & 
        \multicolumn{1}{c}{\makecell{0.506$_{\pm 0.065}$ \\ 0.236$_{\pm 0.072}$}} & 
        \multicolumn{1}{c}{\makecell{0.520$_{\pm 0.057}$ \\ 0.262$_{\pm 0.073}$}} \\
        \midrule
        \makecell{SciBERT \\ \citep{beltagy2019scibert}} & 
        \multicolumn{1}{c}{\makecell{0.098$_{\pm 0.054}$ \\ 0.020$_{\pm 0.021}$}} & 
        \multicolumn{1}{c}{\makecell{0.099$_{\pm 0.075}$ \\ 0.013$_{\pm 0.018}$}} & 
        \multicolumn{1}{c}{\makecell{0.125$_{\pm 0.073}$ \\ 0.047$_{\pm 0.016}$}} &
        \multicolumn{1}{c}{\makecell{0.469$_{\pm 0.112}$ \\ 0.207$_{\pm 0.066}$}} & 
        \multicolumn{1}{c}{\makecell{0.432$_{\pm 0.106}$ \\ 0.183$_{\pm 0.061}$}} & 
        \multicolumn{1}{c}{\makecell{0.446$_{\pm 0.167}$ \\ 0.179$_{\pm 0.071}$}} & 
        \multicolumn{1}{c}{\cellcolor{r1} \makecell{0.481$_{\pm 0.144}$ \\ 0.224$_{\pm 0.010}$}} \\
        \midrule
        \makecell{ScholarBERT \\ \citep{hong2022scholarbert}} & 
        \multicolumn{1}{c}{\makecell{0.286$_{\pm 0.042}$ \\ 0.110$_{\pm 0.009}$}} & 
        \multicolumn{1}{c}{\makecell{0.289$_{\pm 0.044}$ \\ 0.111$_{\pm 0.019}$}} & 
        \multicolumn{1}{c}{\makecell{0.063$_{\pm 0.007}$ \\ 0.005$_{\pm 0.004}$}} &
        \multicolumn{1}{c}{\makecell{0.323$_{\pm 0.058}$ \\ 0.111$_{\pm 0.027}$}} & 
        \multicolumn{1}{c}{\makecell{0.276$_{\pm 0.080}$ \\ 0.076$_{\pm 0.024}$}} & 
        \multicolumn{1}{c}{\cellcolor{r1} \makecell{0.338$_{\pm 0.053}$ \\ 0.117$_{\pm 0.015}$}} & 
        \multicolumn{1}{c}{\makecell{0.296$_{\pm 0.085}$ \\ 0.109$_{\pm 0.044}$}} \\
        \midrule
        \makecell{BioBERT \\ \citep{shoyabiobert}} & 
        \multicolumn{1}{c}{\makecell{0.096$_{\pm 0.171}$ \\ 0.023$_{\pm 0.020}$}} & 
        \multicolumn{1}{c}{\makecell{0.094$_{\pm 0.118}$ \\ 0.015$_{\pm 0.024}$}} & 
        \multicolumn{1}{c}{\makecell{0.042$_{\pm 0.024}$ \\ 0.004$_{\pm 0.001}$}} &
        \multicolumn{1}{c}{\cellcolor{r1} \makecell{0.517$_{\pm 0.031}$ \\ 0.241$_{\pm 0.082}$}} & 
        \multicolumn{1}{c}{\makecell{0.319$_{\pm 0.059}$ \\ 0.110$_{\pm 0.048}$}} & 
        \multicolumn{1}{c}{\makecell{0.424$_{\pm 0.145}$ \\ 0.177$_{\pm 0.119}$}} & 
        \multicolumn{1}{c}{\makecell{0.452$_{\pm 0.114}$ \\ 0.191$_{\pm 0.045}$}} \\
        \midrule
        \makecell{BERT \\ \citep{devlin2018bert}} & 
        \multicolumn{1}{c}{\makecell{0.086$_{\pm 0.032}$ \\ 0.011$_{\pm 0.005}$}} & 
        \multicolumn{1}{c}{\makecell{0.082$_{\pm 0.065}$ \\ 0.012$_{\pm 0.018}$}} & 
        \multicolumn{1}{c}{\makecell{0.034$_{\pm 0.026}$ \\ 0.005$_{\pm 0.006}$}} & 
        \multicolumn{1}{c}{\cellcolor{r1} \makecell{0.566$_{\pm 0.042}$ \\ 0.306$_{\pm 0.073}$}} & 
        \multicolumn{1}{c}{\makecell{0.421$_{\pm 0.137}$ \\ 0.204$_{\pm 0.078}$}} & 
        \multicolumn{1}{c}{\makecell{0.476$_{\pm 0.079}$ \\ 0.225$_{\pm 0.066}$}} & 
        \multicolumn{1}{c}{\makecell{0.520$_{\pm 0.019}$ \\ 0.257$_{\pm 0.022}$}} \\
        \bottomrule
    \end{tabular}
\end{adjustbox}
\end{spacing}
\vspace{-2mm}
\caption{Results of \textbf{slot filling} task among seven tasks on different schema settings for various BERT models pre-trained on different domain specific text data. For each model, the top line represents the micro-F1 score and the bottom line represents the macro-F1 score. We report the mean across 5 experiments with a confidence interval of two standard deviations. We highlight the \hlc[r1]{best} performing method.}
\label{table:schema-sf}
\end{table*}

\end{document}